\documentclass[twoside,10pt]{article}
\PassOptionsToPackage{square, numbers}{natbib} 
\usepackage{jmlr2e}
\usepackage[utf8]{inputenc} 
\usepackage[T1]{fontenc}    

\usepackage{hyperref}       
\usepackage{url}            
\usepackage{booktabs}       
\usepackage{amsmath}
\usepackage{amssymb}
\usepackage{amsfonts}
\usepackage{amsthm}         
\usepackage{bm}
\usepackage{mathtools}
\usepackage{thmtools}
\usepackage{thm-restate}
\usepackage{nicefrac}       
\usepackage{microtype}      
\usepackage{wrapfig}
\usepackage{multirow}
\usepackage{adjustbox}
\usepackage{natbib}
\usepackage{cleveref}
\usepackage{enumitem}
\usepackage{xcolor}
\usepackage{xspace}
\usepackage{subcaption}
\usepackage{stmaryrd}
\usepackage{float}
\usepackage[version=4]{mhchem}
\usepackage[frak=pxtx]{mathalpha}
\usepackage{mathrsfs}
\usepackage{algorithm}
\usepackage{algorithmic}
\usepackage{diagbox}
\usepackage{comment}
\usepackage{tikz}
\usetikzlibrary{calc,patterns,angles,quotes}

\Crefname{equation}{Eq.}{Eqs.}
\Crefname{figure}{Fig.}{Fig.}
\Crefname{tabular}{Tab.}{Tabs.}
\Crefname{table}{Tab.}{Tabs.}
\Crefname{section}{Sec.}{Sec.}
\Crefname{appendix}{App.}{App.}

\akbcheading
\ShortHeadings{3D detection of roof sections at LOD2 by KIBS}{Lussange, Yu, Tarabalka, Lafarge}
\finalcopy

\begin{document}

\title{\vspace{-9mm}3D detection of roof sections from a single satellite image and application to LOD2-building reconstruction}

\author{\name Johann Lussange$^{1}$ \email johann.lussange@inria.fr \\
       \name Mulin Yu$^{1}$ \email mulin.yu@inria.fr \\
       \name Yuliya Tarabalka$^{2}$  \email ytarabalka@luxcarta.com \\
       \name Florent Lafarge$^{1}$ \email florent.lafarge@inria.fr \\
       \\
      \addr $^{1}$ INRIA Sophia Antipolis Méditerrannée, 2004 route des Lucioles, 06902, Valbonne, France. \\
      \addr $^{2}$ LuxCarta Technology, 460 avenue de la Quiéra, voie K, bat. 119 B, 06370, Mouans-Sartoux, France. }
\maketitle

\begin{abstract}

Reconstructing urban areas in 3D out of satellite raster images has been a long-standing and challenging goal of both academical and industrial research. The rare methods today achieving this objective at a Level Of Details $2$ 
rely on procedural approaches based on geometry, and need stereo images and/or LIDAR data as input. We here propose a method for urban 3D reconstruction named KIBS(\textit{Keypoints Inference By Segmentation}), which comprises two novel features: i- a full deep learning approach for the 3D detection of the roof sections, and ii- only one single (non-orthogonal) satellite raster image as model input. This is achieved in two steps: i- by a Mask R-CNN model performing a 2D segmentation of the buildings' roof sections, and after blending these latter segmented pixels within the RGB satellite raster image, ii- by another identical Mask R-CNN model inferring the heights-to-ground of the roof sections' corners via panoptic segmentation, unto full 3D reconstruction of the buildings and city. We demonstrate the potential of the KIBS method by reconstructing different urban areas in a few minutes, with a Jaccard index for the 2D segmentation of individual roof sections of $88.55\%$ and $75.21\%$ on our two data sets resp., and a height's mean error of such correctly segmented pixels for the 3D reconstruction of $1.60$ m and $2.06$ m on our two data sets resp., hence within the LOD2 precision range. 

\end{abstract}

\section{Introduction} 
\label{SectionI}

In the rapidly evolving era of smart cities and intelligent urbanization, digital city models have become crucial tools for urban planning, environmental analysis, and infrastructure management. Relying on satellite, aerial, and \textit{LIght Detection and Ranging} (LIDAR) imagery, these models offer detailed three-dimensional representations of urban environments, and facilitate better-informed decision-making processes. At the same time, computer vision research in satellite and aerial imagery has made great strides in recent years. However, the unique challenges posed by satellite, aerial, and LIDAR imagery, such as variation in perspective, scale, lighting, atmospheric conditions, and data density, necessitate constant technical advancements. New algorithms and models have allowed for much more accurate and efficient image analysis, notably with the rise of deep learning methods. In a larger scope, these have shown much promise in automatically detecting and classifying objects in images~\cite{Zhu2017,Jia2019,Maggiori2016}. This object detection and classification is especially relevant to the fields of semantic segmentation~\cite{Bahl2022} and 3D reconstruction~\cite{Ji2019}. With the ever-increasing availability of data (notably LIDAR data~\cite{Bauchet2019,Zhu2017}), new applications are constantly arising and allow for the automation of detection and correction of various types of distortion in images, such as those caused by atmospheric conditions~\cite{Bahlb2022} and the curvature of the Earth's surface~\cite{Dongsung2020}, or building~\cite{Liuyun2016,Girard2021,Maggiori2017} and vegetation occlusion~\cite{Vali2020}, etc. Also, new methods are being developed for automatically extracting information from raster images~\cite{Ghamisi2018} such as land cover~\cite{Hosseiny2022}, or topographical features~\cite{Bourque2018}. In this Paper, we will first give a brief overview of the related work and other pertaining methods in Section \ref{SectionII}. We will then proceed to explain our own proposed KIBS (\textit{Keypoints Inference By Segmentation}) approach in Section \ref{SectionIII}, where we will describe the model's two-steps architecture and its data post-processing. In Section \ref{SectionIV}, we will then describe our experiment, with a presentation and discussion of the results of the method on our data set, together with details on the model generalisation and limitations. We also describe the training, validation and testing procedure for Mourmelon-le-Grand and Sissonne data sets, for both the 2D segmentation and the 3D reconstruction of the KIBS method in the Section \ref{SectionVI} of the Supplementary Material.

\section{Related work} 
\label{SectionII}

\begin{figure}[!htbp]
\begin{centering}
\includegraphics[scale=0.062]{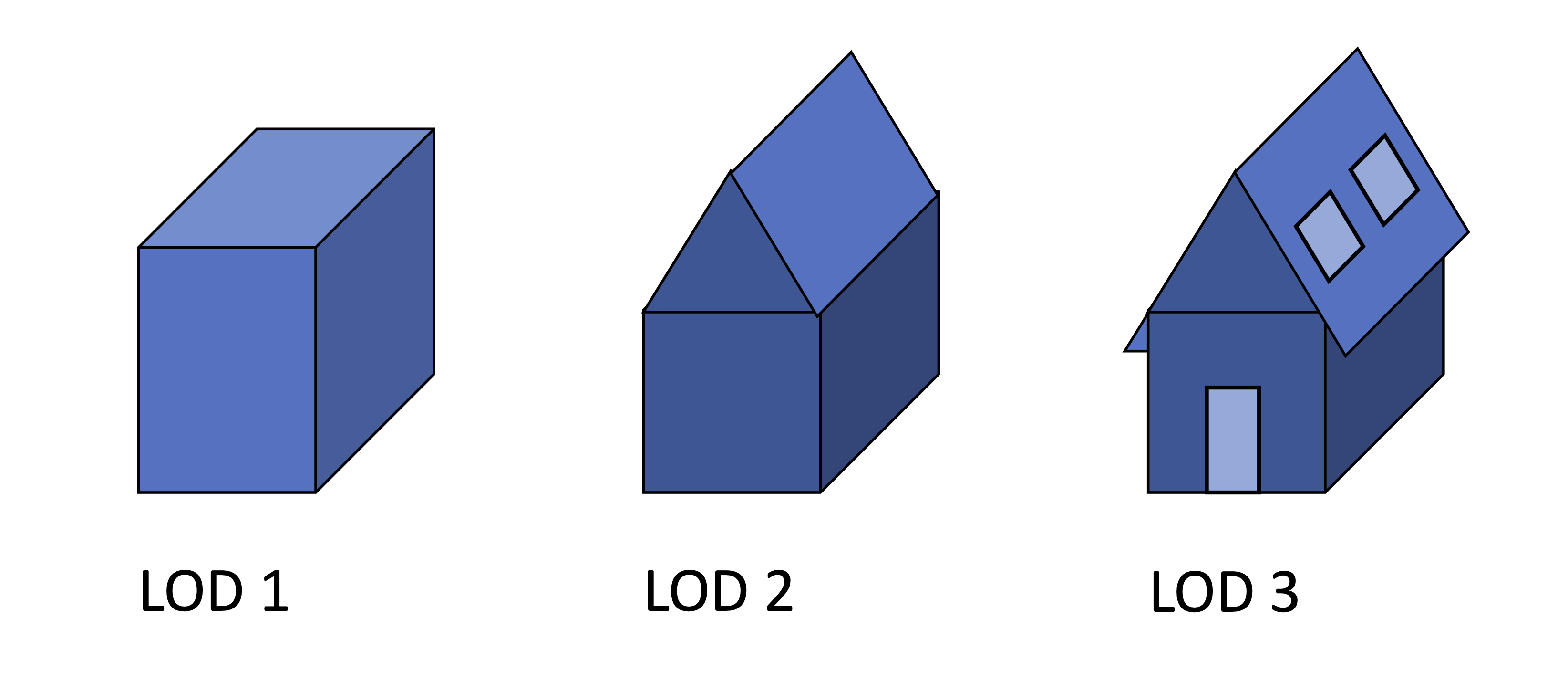}
\includegraphics[scale=1.04]{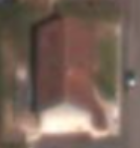}
\caption{\label{A1} Left: examples of Level of Details (LOD) of 3D reconstructions. Right: random residential building in our satellite image input data set of Mourmelon-le-Grand, which has a precision of $0.38$ meter per pixel, to compare its precision with that of the 3D reconstruction goal at a LOD2 (cf. car parked on the right-hand side of the house).}
\end{centering}
\end{figure}

The interest of using satellite data as input for reconstruction relies on the abundance and low costs of such data, compared to other sources such as LIDAR or aerial data, which face legal or technical constraints, flight authorisation issues over certain areas, etc. The Level Of Details (or LOD) is a usual metric that allows one to specify the desired precision of such reconstruction. As shown on Fig. \ref{A1}, LOD1 denotes a building reconstruction precision looking like a rectangular shoe box, LOD2 denotes a reconstruction precision displaying the shape of the building's roof, while LOD3 denotes a reconstruction precision of objects' sizes below this range, such as windows, balconies, etc. A few studies have claimed to perform urban 3D reconstruction at a LOD1~\cite{Tripodi2019}, or similar outcomes of ground surface reconstructions, but the portability of such methods have often remained limited due to their highly procedural architectures~\cite{Leotta2019}. A method for 3D reconstruction at LOD2 is being patented~\cite{patentluxcarta}, but proposes a very different approach than the one-shot procedure presented here, by using pre-existing primes of rooftops. Most of such methods also rely on extra data sets that are not purely of satellite origin~\cite{Zhao2022,Kudinov2021}, such as LIDAR data~\cite{Chatterjee2019,Li2019}, aerial photography~\cite{Yu2021}, etc. Others rely on data sets of pre-existing primes of rooftops~\cite{Ren2021,patentluxcarta}. We here review the latest advancements in 3D plane detection and reconstruction, which is an active area of research within computer vision, with substantial contributions made through the use of both single-image and multi-view images or point cloud data. 

\paragraph{Plane detection from single image}
Single-image plane detection and reconstruction have seen remarkable progress thanks to advancements in deep learning. Researchers have developed several methods to detect and reconstruct planes using just a single image. For example, in the PlaneFormers paper \cite{PlaneFormers22}, they utilize deep learning to develop an algorithm that can reconstruct 3D planes from sparse view planes. Another method, PlaneRCNN, was proposed by \cite{PlaneRCNN19} that detects and reconstructs 3D planes from a single image using a Region Convolutional Neural Network. Similarly, \cite{Yu2019} proposed a method for single-image piece-wise planar 3D reconstruction via associative embedding, and \cite{PlaneNet18} introduced PlaneNet for piece-wise planar reconstruction from a single RGB image. Further, \cite{yang_eccv18} employed convolutional neural networks for recovering 3D planes from a single image, highlighting the potential of deep learning in plane detection and reconstruction from single images. However these methods are designed to extract a few large planes in certain types of images, typically indoor scenes, and fails to detect the numerous small planes, e.g. hundred of thousand, contained in a satellite image representing a city.

\paragraph{Plane detection from point clouds and multiview images}
Moving beyond single images, point clouds and multiview images offer additional information that can be leveraged for plane detection and reconstruction. Classical methods such as region growing \cite{rabbani2006segmentation,vo2015octree} and RANSAC \cite{schnabel2007efficient,mordohai19} have been widely used for this purpose. On the other hand, energy minimization methods \cite{monszpart2015rapter,Yu2022} provide a more rigorous approach, leveraging the mathematical foundation of energy functions for plane detection. Scale-space exploration, as demonstrated by \cite{fang_cvpr18,mellado_20}, is another valuable technique that adapts to various scales for improved detection. Recently, deep learning-based methods have shown great promise, offering new opportunities for plane detection from point clouds and multiview images \cite{li2019supervised,parsenet_eccv20,le2021cpfn,primitivenet_iccv21,HPNet_iccv21}. Unfortunately, such techniques cannot be used in our context where point clouds generated by MVS from satellite imagery have a very low precision on the spatial coordinates of points.  

\paragraph{Building reconstruction}
Roof reconstruction has been a challenging task in 3D building modeling, requiring special attention. Different data sources provide different opportunities and challenges for roof reconstruction. In this context, roof skeletonization techniques \cite{Ren2021} and deep learning-based aerial image analysis methods \cite{Yu2021,Zhao2022} have shown promising results. In the case of LiDAR data, methods like \cite{Bauchet2019} have proven effective. Generative models like Roof-GAN \cite{Roof-GAN21} have demonstrated the ability to learn and generate roof geometry and relations for residential houses. In another approach, \cite{zeng_eccv18} proposed neural procedural reconstruction for residential buildings, merging the power of deep learning with the procedural generation approach. Such approaches operate from aerial data and are not robust anymore from satellite data. 

Reconstructing urban environments in 3D out of satellite raster images has been a long-standing ambitious objective of both industrial and academic research~\cite{Goldberg2017}. One exciting application of plane detection and reconstruction is in building reconstruction from satellite images. Researchers have employed a variety of methods for this task. For instance, automated building extraction from satellite imagery is explored in \cite{gui2021automated}, and building detection from remotely sensed data is studied in \cite{rs11141660}. In the context of LOD2 models, roof type classification plays a crucial role, as seen in \cite{Leotta2019}, which utilized PointNet for this purpose. For LOD1 models, beyond \cite{Tripodi2019}, methods like Voronoi-based algorithms \cite{Liuyun2016} and polygonalization of footprints \cite{Girard2021,Zorzi_2022_CVPR,polymapper_2019} have shown effectiveness. Furthermore, advancements in dense mesh and Digital Surface Model (DSM) generation techniques, such as IMPLICITY, which uses deep implicit occupancy fields for city modeling from satellite images \cite{stucker2022implicity}, have further pushed the boundaries of what is possible in this domain.

To the best of our knowledge, we are the first to try reconstruct LOD2-building by detecting and assembling planes directly from one single satellite image.

\section{Proposed approach} 
\label{SectionIII}

\subsection{Overview}
\label{SectionIIIa}

The KIBS procedure performs the 3D reconstruction of urban areas at a LOD2 with, compared to previous methods, two new features: i- a full deep learning solution for the 3D detection of the buildings' roof sections, and ii- an input consisting in only one single satellite raster image. In order to do this, the KIBS model follows a two-step procedure: first a 2D segmentation task identifies the roof sections, and then a second 3D reconstruction task infers those roof sections' corners with their height-to-ground (as a unique class). Such monocular or single-view 3D reconstruction approaches have been recently used in the general field of computer vision \cite{Choy2016,Chen2019,Fu2021}, however, these methods were applied to simpler images (like those of individual objects or indoor scenes), and applying them to complex raster data like satellite imagery of urban areas is a more challenging problem. 

\paragraph{Input} The KIBS method is here trained on a data set of satellite raster images with a precision of $0.38$ meter per pixel (see Fig. \ref{A1} for a sample). The input of the first data set used for the training, validation, and testing of this method derives from one RGB satellite image of Mourmelon-le-Grand, France, of size $30564 \times 26320$ pixels, corresponding to a surface area of $\sim 73$km$^2$. This raster image comes from Maxar's Worldview-3 satellite, and was acquired on the $13$th of August $2020$, with a satellite azimuth angle of $181.10^{\circ}$, elevation angle of $59.30^{\circ}$. This raster image is accompanied with a data set serving as ground truth for this outcome of urban 3D reconstruction, that consists in a hand-annotated shapefile of all individual roof sections' contours, together with their corners' heights above mean sea level. It is also accompanied with a Digital Terrain Model (DTM, i.e. an elevation map of the ground surface, without its urban or natural objects), courtesy of LuxCarta. Once the KIBS method has been developed for Mourmelon-le-Grand, the model has been trained, validated and tested on a second, similar, data set in order to further confirm its validity, this time on the city of Sissonne, France, whose raster data is of size $19120 \times 17420$, corresponding to a surface area of $\sim 25$km$^2$. This raster image also comes from Maxar's Worldview-3 satellite, and was acquired on the $4$th of November $2020$, with a satellite azimuth angle of $172.9^{\circ}$, elevation angle of $66.6^{\circ}$. It also comes with a DTM specifying the altitudes to sea-level of the terrain, and a hand-annotated shapefile of all individual roof sections' contours, together with their corners' altitudes to sea-level. 

\begin{figure}[!htbp]
\begin{centering}
\includegraphics[scale=0.1]{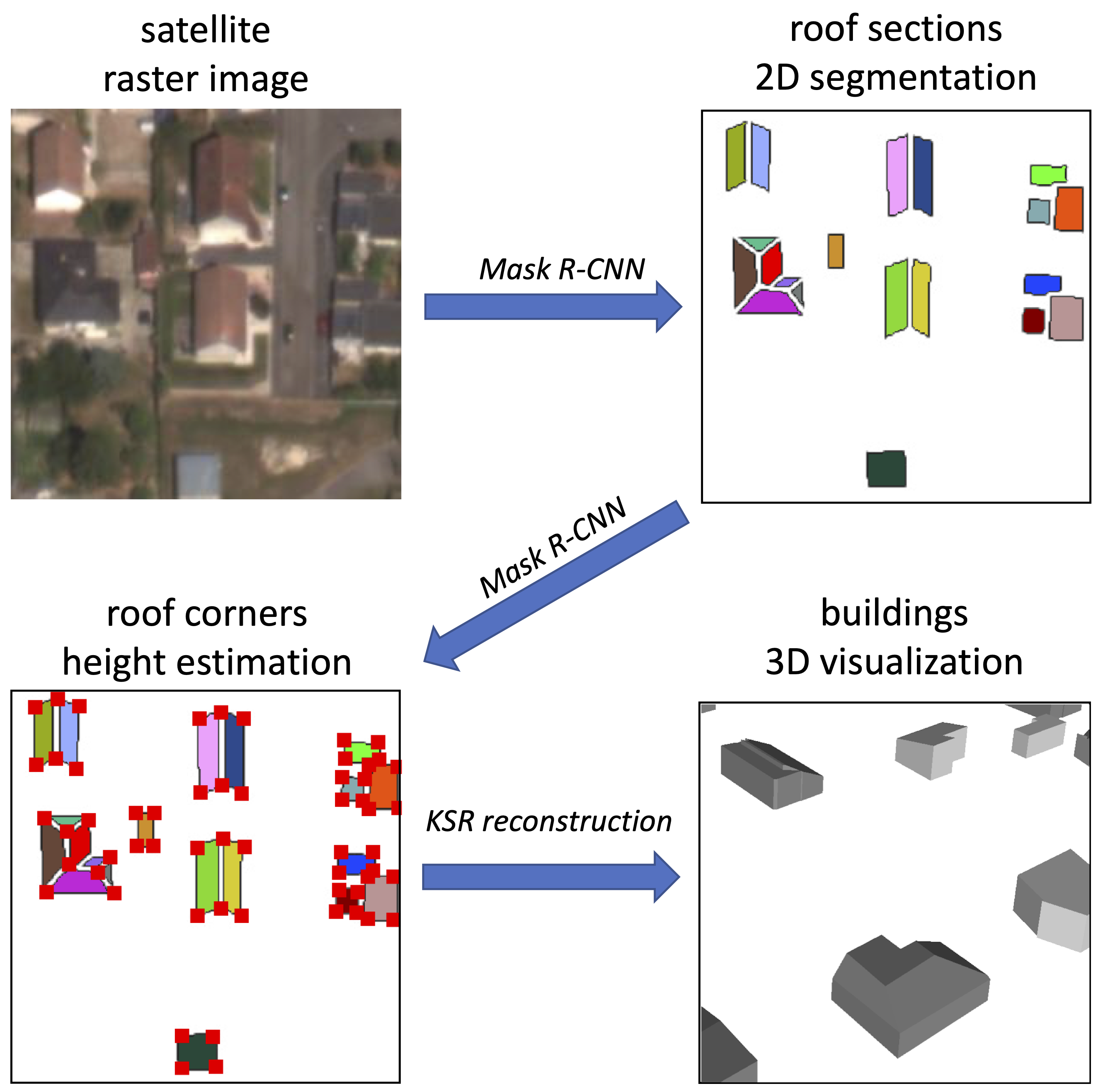}
\caption{\label{A2} General procedure of the KIBS method. A first Mask R-CNN model takes the satellite raster image as input and performs an individual 2D segmentation of the buildings' roof section. Then each segmented pixel of this output is blended back into that same RGB raster image, and serves as input to a second, distinct, Mask R-CNN model in order to both identify the roof corners, and estimate their height-to-ground (as a class in meters, according to the LOD2 precision standards, i.e. $1$ m, $2$ m, $3$ m, etc.). The inference of at least three roof corners allows one then to derive the associated roof section plane 3D coefficients, and hence recover the whole building 3D reconstruction, that one can then visualise with the KSR geometrical procedures.}
\end{centering}
\end{figure}

\paragraph{Output} Once trained, the first part of the KIBS model outputs a 2D segmentation of the roof sections, which is fed into a second part of the model employing panoptic segmentation in order to derive those roof section corners' height-to-ground, so as to compute their associated 3D planes coefficients, unto full building and urban area reconstruction. We finally use the Kinetic Shape Reconstruction (KSR) method developed in~\cite{Yu2022} in order to visualize it. A sketch of the whole KIBS procedure is shown in Fig. \ref{A2}. 

\paragraph{Hypotheses} The general working hypothesis of this research study is that it is possible to perform the 3D reconstruction of buildings at a LOD2, for a model taking as input only one single, non-orthogonal, satellite raster image with a resolution of $0.38$ meter per pixel (see Fig. \ref{A1} for a comparison). More specifically, within the scope of the KIBS method, our working hypothesis is that a deep learning approach can segment in 2D and reconstruct in 3D the roof sections of the buildings of an urban area with a LOD2, at this image resolution, and based on a single-shot satellite raster image. The fundamental intuition behind this hypothesis is that the non-orthogonality of the satellite raster image provides the deep learning algorithms with non-trivial information (e.g. buildings' walls' inclination, buildings' shadows, roof peak or ridge perspective, etc.) allowing them to infer the height-to-ground of the roof sections' corners with a precision within the bounds of the LOD2 requirement.

\subsection{2D detection of roof sections}
\label{SectionIIIb}

The process for training data preprocessing for the Mask-RCNN model for 2D segmentation of roof lines involves several steps. Firstly, the initial $8687 \times 9890$ satellite image is segmented into individual $230 \times 230$ tiles, overlapping by a margin of $10$ pixels. Subsequently, ground truth shapefile polygons delimiting roof sections are extracted from these tiles. Each tile then gets a set of corresponding black and white images with white pixels representing a unique roof section per image. 

Using the \verb?pycococreator? algorithm \cite{pycococreatorgithub}, annotation files in PYCOCO format are created for these ground truth masks. After randomly shuffling the set of tiles and associated ground truth images, it is divided into three disjoint sets: training ($60\%$), validation ($20\%$), and testing ($20\%$).

These sets and their associated annotation files are then fed into a Mask-RCNN neural network named \verb?mask_rcnn_R_50_FPN_3x?, a combination of a ResNet-50 model stacked with a Feature Pyramid Network (FPN). This model was chosen due to its robustness and ability to handle complex segmentation tasks. The training, which ran for six days on a Dell T630 GPU node with four GeForce GTX 1080 Ti GPUs, was monitored via TensorBoard to manage regularization issues. The trained network weights are available on the KIBS GitHub repository. More implementation details on the training procedure, as well as the training metrics are given in Section \ref{SectionVIa} and \ref{SectionVIc} of the Supplementary Material, respectively. The weights of the trained model, which represent the learned features, are available on the KIBS GitHub repository for further exploration and reproducibility of our results \cite{biyologithub}.

\subsection{3D roof corners extraction}
\label{SectionIIIc}

The 3D reconstruction training leverages a Mask-RCNN model, similar to the 2D segmentation process but geared towards panoptic segmentation. This involves marking roof section corners on the image output of the 2D segmentation and assigning unique class labels to these corners, representing their heights.

After training, the 2D segmentation output is integrated with the original RGB raster image, improving the 3D reconstruction's efficiency in identifying roof corners. Class labels corresponding to specific heights are used in the Detectron2 framework, extendable to handle taller structures.

Generating the training, validation, and testing sets follows a similar procedure to the 2D segmentation. Each blended raster image is linked with ground truth images representing roof corners, and this data is processed via pycococreator to create annotation files compatible with the Detectron2 framework. A Mask-RCNN model is trained to recognize roof corners and their heights. The training process, monitored online to manage regularization issues, leverages the same hardware as the 2D segmentation, with model weights available on the KIBS GitHub repository. More implementation details on the training procedure, as well as the training metrics are given in Section \ref{SectionVIb} and \ref{SectionVId} of the Supplementary Material, respectively.

\subsection{Plane estimation and meshing}
\label{SectionIIId}

As said, once at least three roof section's corners are inferred, and their height-to-ground estimated, one can easily geometrically derive the 3D plane coefficients of the associated roof section, and hence the height-to-ground of each pixel belonging to this roof section, unto full building and then city-wide 3D reconstruction. Now for a number $N \geqslant 4$ of segmented roof section corners, the algorithm proceeds to select three corners among these forming the largest triangle area via a basic Delaunay triangulation, so as to increase 3D reconstruction accuracy, as shown in Fig. \ref{A2}. 

\subsection{Implementation details}
\label{SectionIIIe}

We can cover the testing procedure of the KIBS method in five general steps. 

Firstly, the whole satellite raster image is split in a grid of $230 \times 230$ tile images, with a margin overlap of $10$ pixels on each four sides of the image. 

Secondly, the aforementioned Mask-RCNN model trained for 2D segmentation is applied to each of these tile images so as to infer the roof sections 2D segmentation. 

Thirdly, these segmented pixels are blended within their associated raster tile image as blue pixels, with a value $\{0, 0, 200 \}$ if belonging to the training data set, $\{0, 0, 210 \}$ if belonging to the validation data set, and $\{0, 0, 220 \}$ if belonging to the testing data set. 

Fourthly, the aforementioned Mask-RCNN model trained for 3D reconstruction is then applied to each of these blended tile images so as to infer the roof section corners as keypoints, with their own height-to-ground (as a class in meters, according to the LOD2 precision standards, i.e. $1$ m, $2$ m, $3$ m, etc.). After some postprocessing, the output represents these roof section squares as red squares of $15 \times 15$ pixels where the red RGB channel is given the value $200 + z$, where $z \in \mathbb N^{\ast}$ is the height-to-ground of the corner, as shown on Fig. \ref{B2a} for Mourmelon-le-Grand and \ref{B2b} for Sissonne. 

Fifthly, as already said, for $N \geqslant 3$, one can easily geometrically derive the 3D plane coefficients of the roof section, and hence the height-to-ground of each pixel belonging to this roof section, unto full building and then city-wide 3D reconstruction. The latter can then be visualised in 3D via the Kinetic Shape Reconstruction (KSR) method developed in~\cite{Yu2022} (see below). 

\begin{figure}[htbp]
\begin{centering}
\includegraphics[scale=0.065]{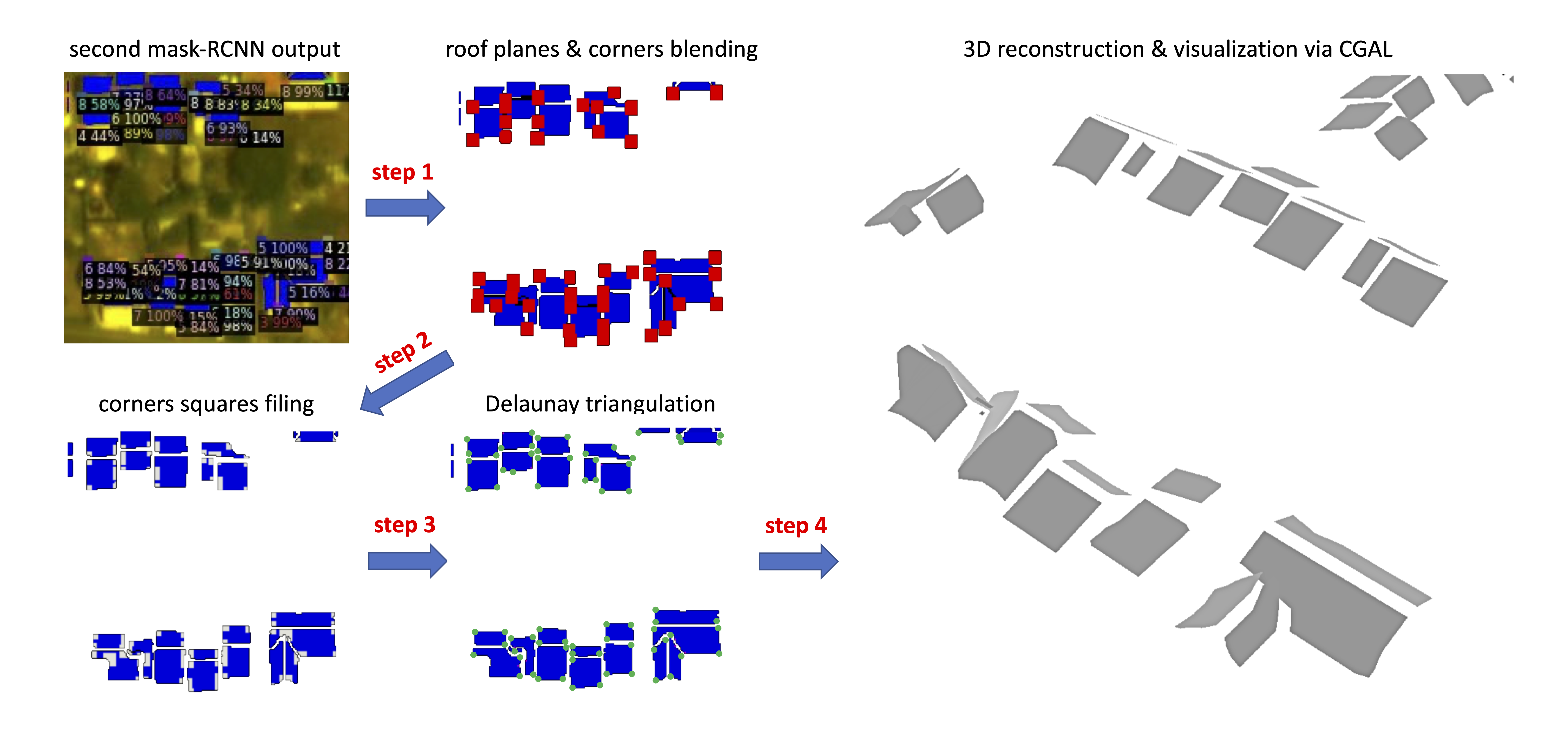}
\caption{\label{A3} Step by step data postprocessing pipeline of the 3D reconstruction output. Firstly, the code translates the raw output of the 3D reconstruction Mask-RCNN algorithm into a blending of the roof sections from the first algorithm (blue pixels) with their corners (red pixels). Then in a second step, the red $15 \times 15$ roof section squares are filed so as to remain on their associated roof sections only, and not on neighbouring ones, nor outside of any plane structure at all. In a third step, the code determines by Delaunay triangulation which are the three roof corner pixels forming the largest possible triangle area for each roof section (when possible), so as to derive its 3D plane coefficients in a more precise manner. Eventually, the code can use the Computational Geometry Algorithms Library~\cite{cgal} (CGAL) in order to visualize the basic roof sections' reconstruction in 3D.}
\end{centering}
\end{figure}

The details of the data postprocessing of the KIBS model are given in the section \ref{SectionVIe} of the Supplementary material. We here simply sum up this procedure via Fig. \ref{A3} as a general description. 

\section{Experiments}
\label{SectionIV}

\subsection{Qualitative results} 
\label{SectionIVa}

\begin{figure}[htbp]
\begin{centering}
\includegraphics[scale=0.6]{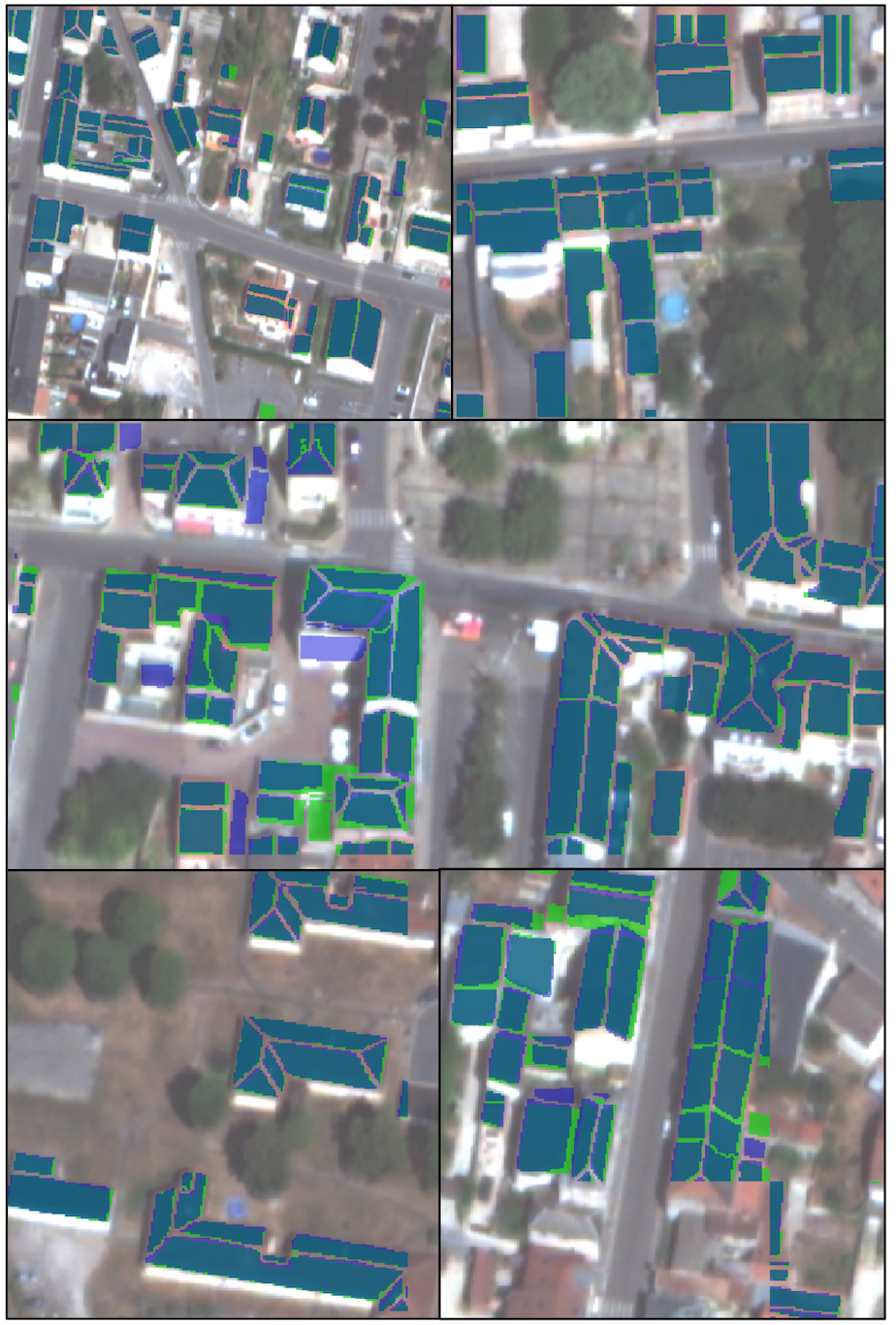}
\caption{\label{B1a} 2D segmentation output visualization of Mourmelon-le-Grand (on the testing set only, which consists in shuffled $230 \times 230$ tiles from the whole raster data), with the ground truth being displayed in light green and the model output in blue. The model predicts segmented pixels correctly when both overlap, in dark green.}
\end{centering}
\end{figure}

\begin{figure}[htbp]
\begin{centering}
\includegraphics[scale=0.6]{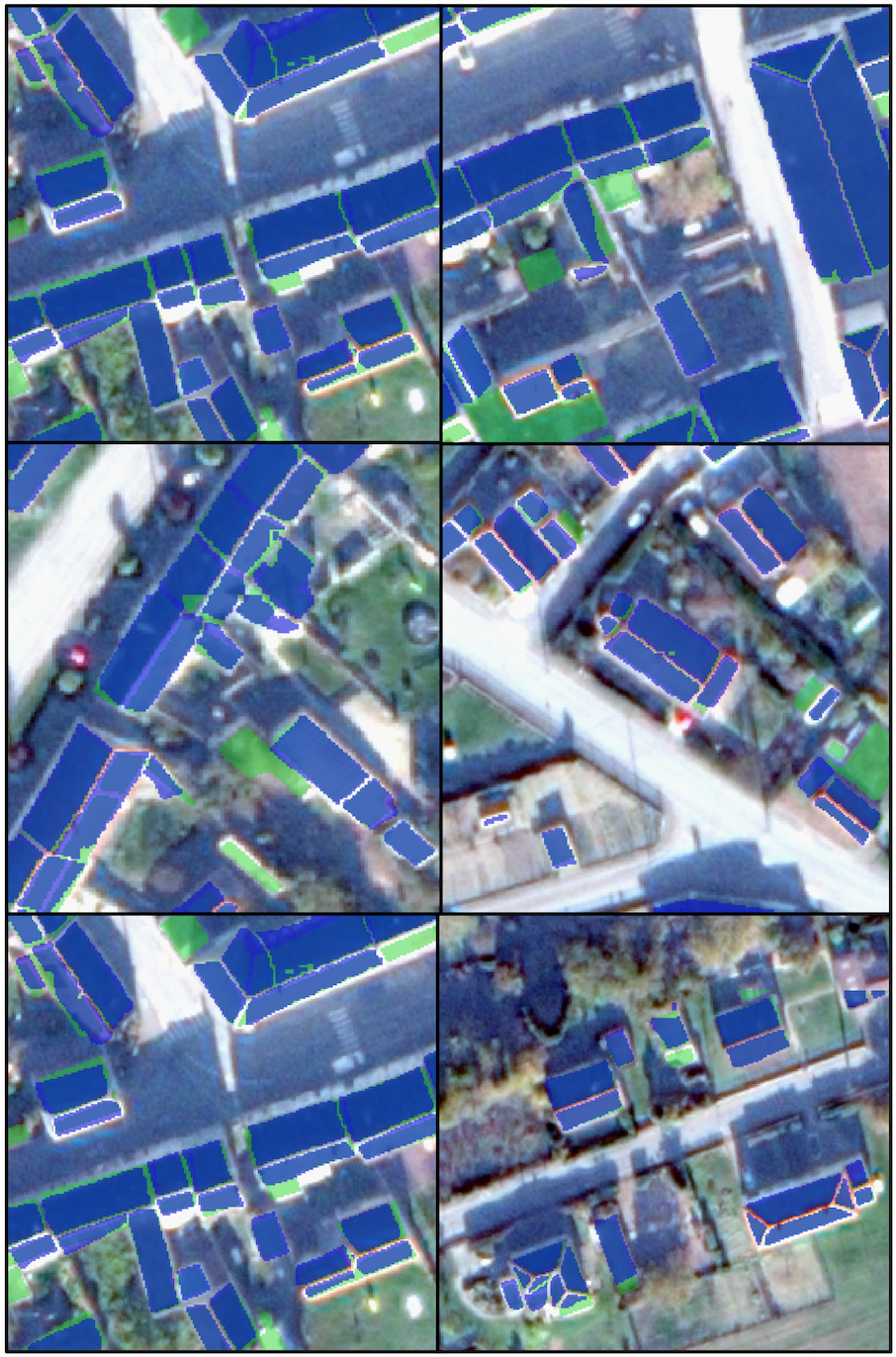}
\caption{\label{B1b} 2D segmentation output visualization of Sissonne (on the testing set only, which consists in shuffled $230 \times 230$ tiles from the whole raster data), with the ground truth being displayed in light green and the model output in blue. The model predicts segmented pixels correctly when both overlap, in dark green.}
\end{centering}
\end{figure}

The results of the 2D segmentation of the roof sections for all data sets (training, validation, testing) are shown in Fig. \ref{B1a} for Mourmelon-le-Grand and Fig. \ref{B1b} for Sissonne. These figures provide a detailed visual comparison between the original satellite images and the output of the 2D segmentation part of the KIBS model, allowing to qualitatively assess the accuracy and precision of our model in identifying and segmenting the roof sections from the satellite images. One can thus see the model's capability to accurately perform 2D segmentation of urban satellite images, which is a crucial step towards achieving our ultimate goal of 3D urban reconstruction.

\begin{figure}[htbp]
\begin{centering}
\includegraphics[scale=0.1565]{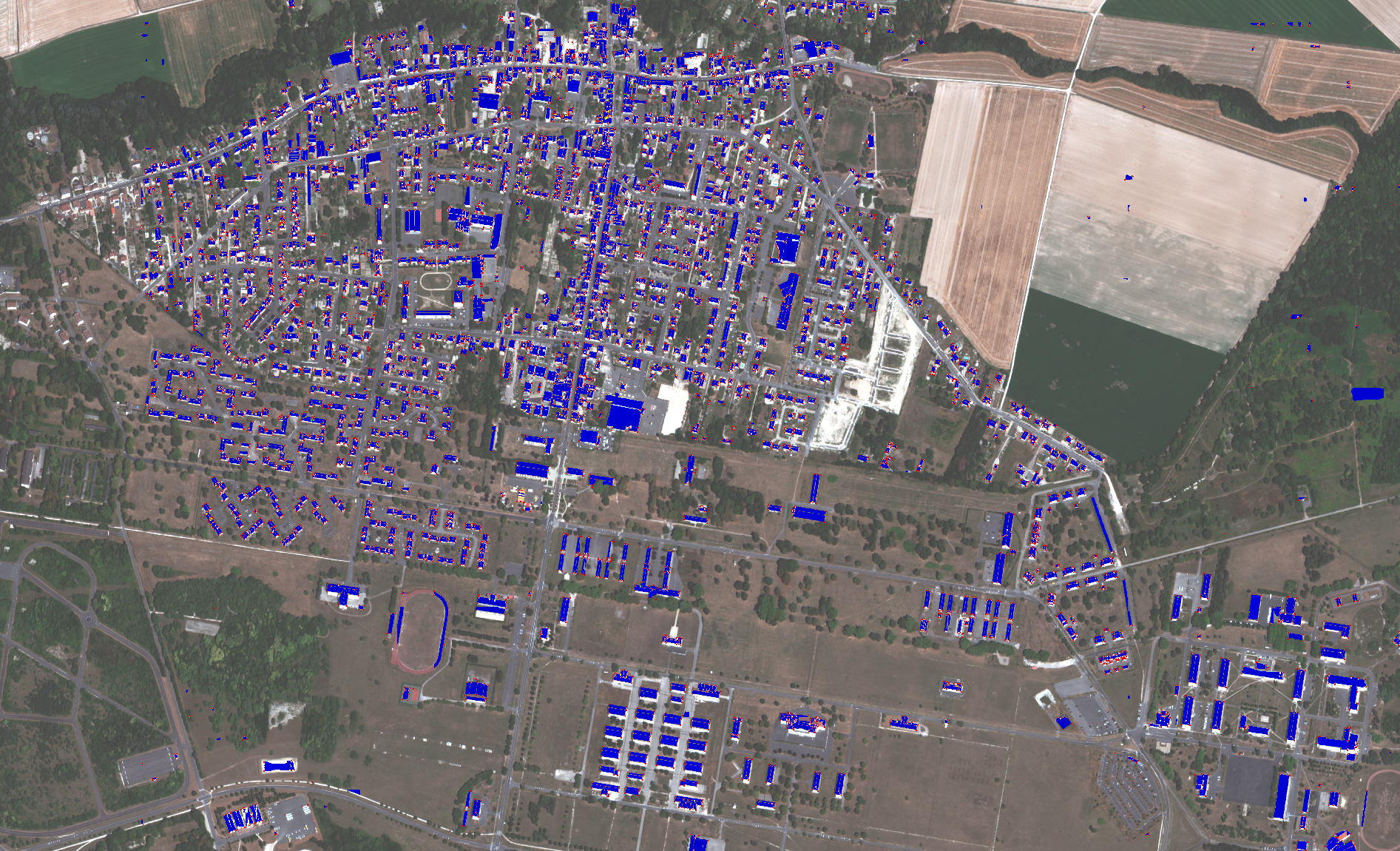}\\
\includegraphics[scale=0.14]{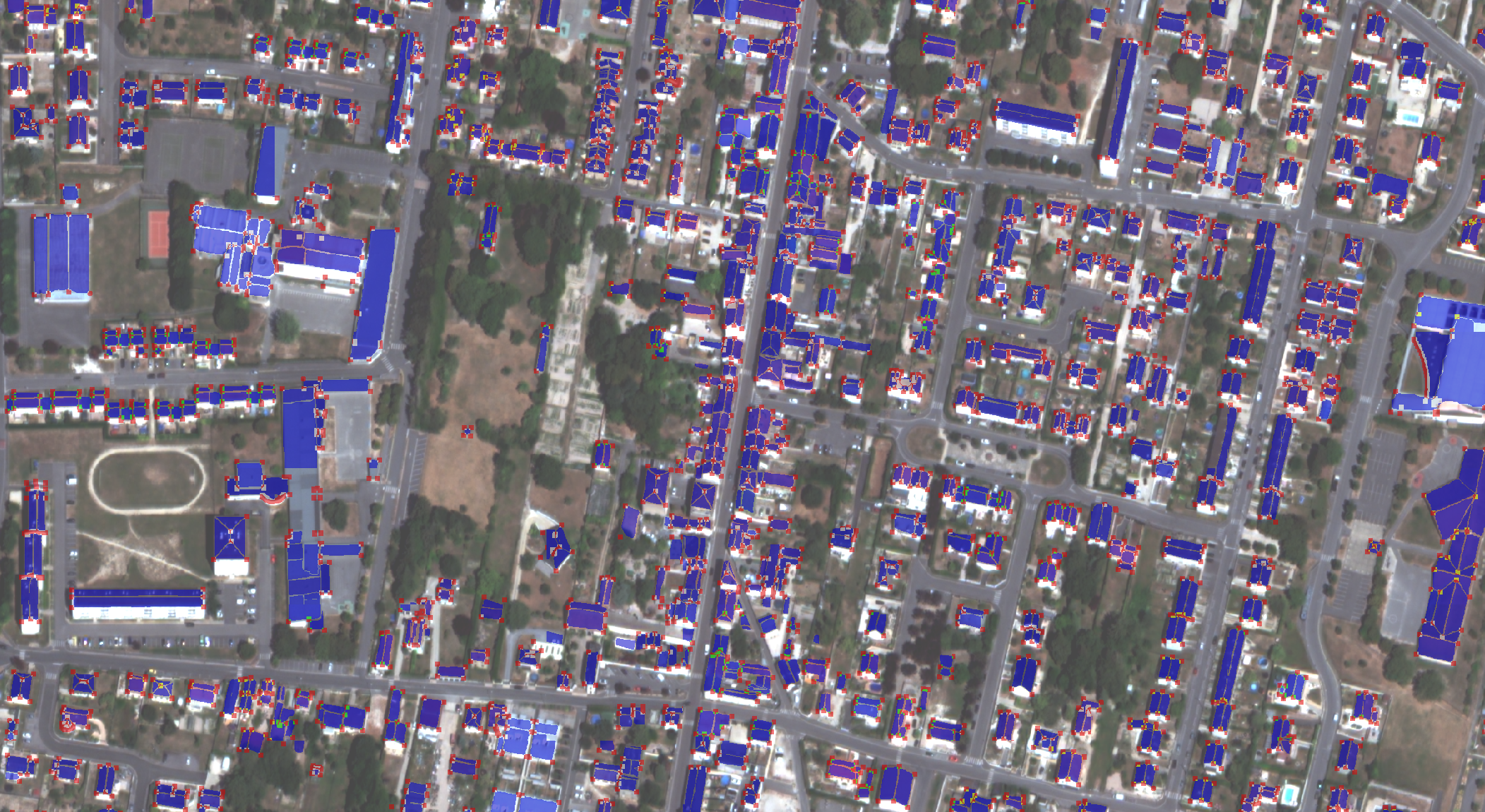}\\
\includegraphics[scale=0.14]{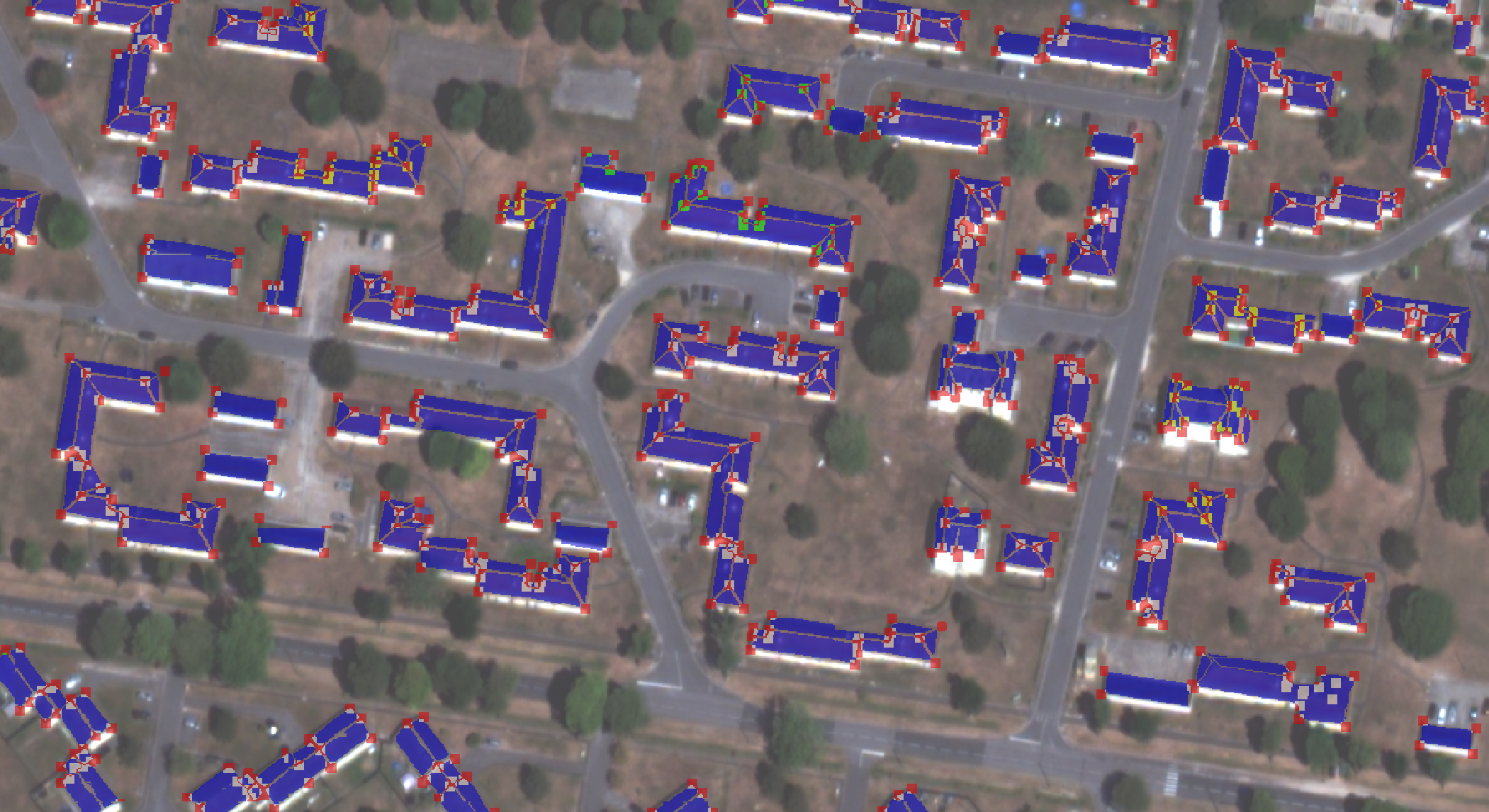}
\caption{\label{B2a} Outputs of the roof section keypoints inference before data postprocessing, overlaying the associated raster image of Mourmelon-le-Grand. Each red square contains the height-to-ground $z$ of its associated roof corner within the red channel of the RGB picture, with a value $200+z$. For visualization purposes, when these overlap the blue segmented pixels of their associated roof section(s), they take a white, green, or yellow color if they belong to the training, validation, or testing set, respectively.}
\end{centering}
\end{figure}

\begin{figure}[htbp]
\begin{centering}
\includegraphics[scale=0.15]{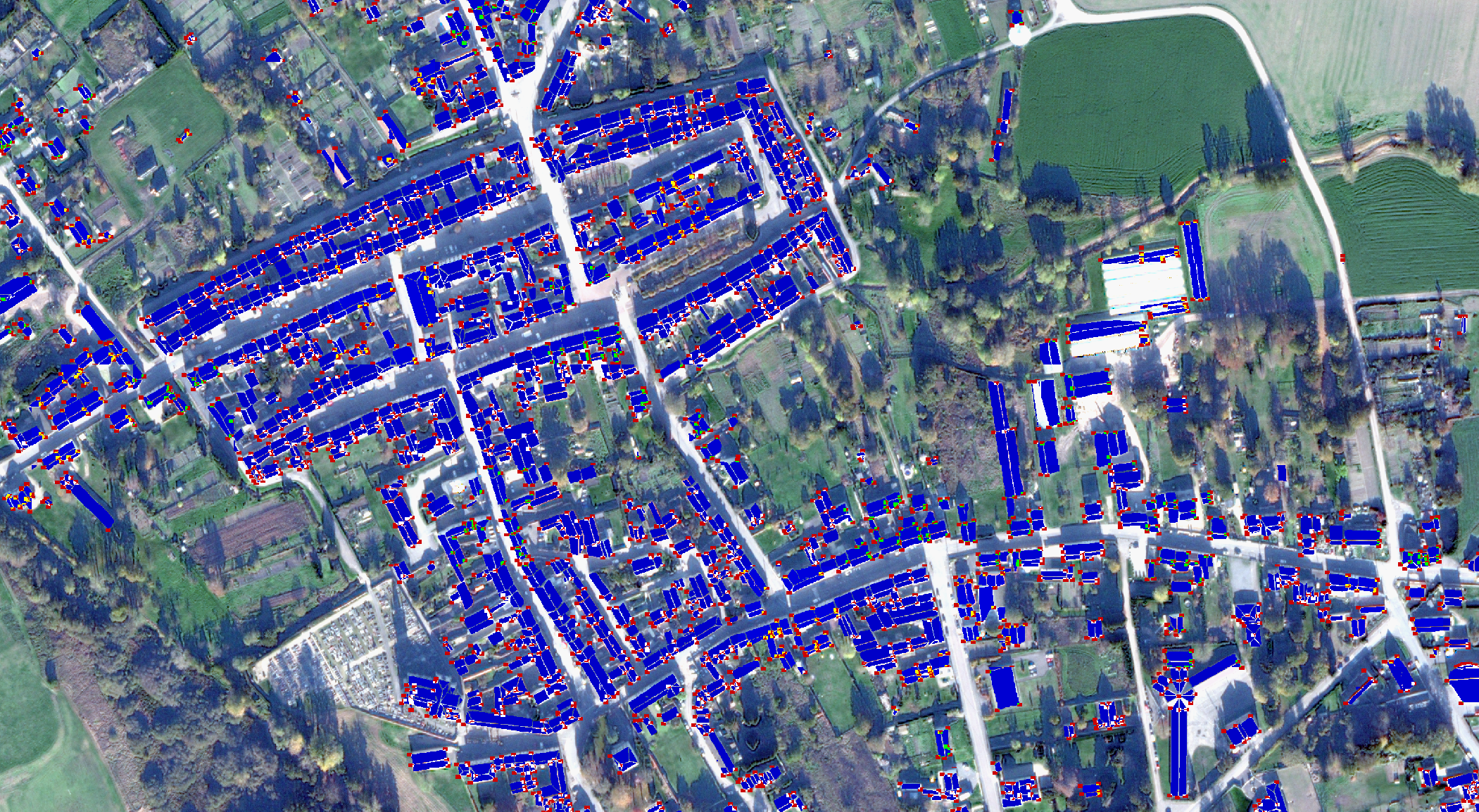}\\
\includegraphics[scale=0.15]{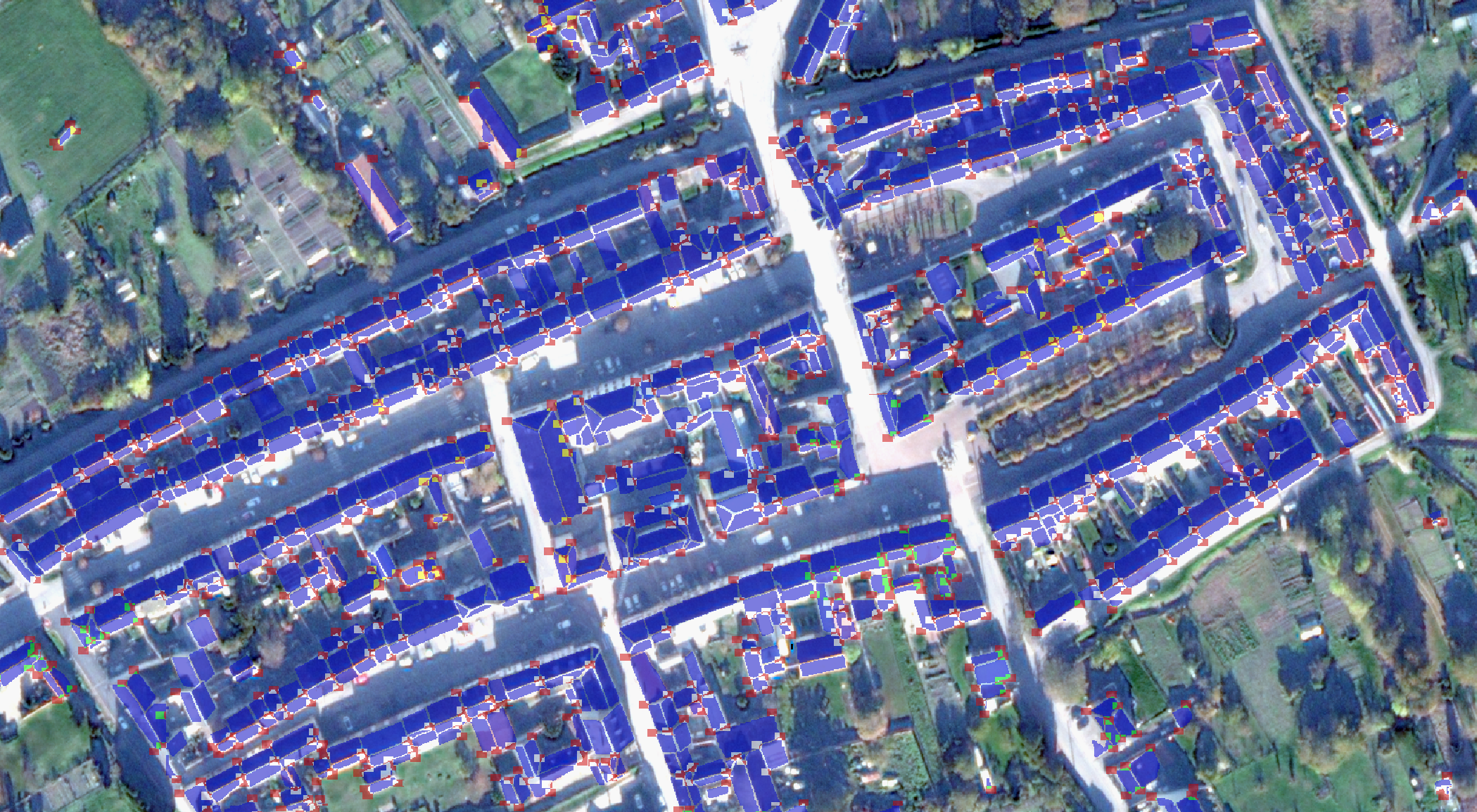}\\
\includegraphics[scale=0.15]{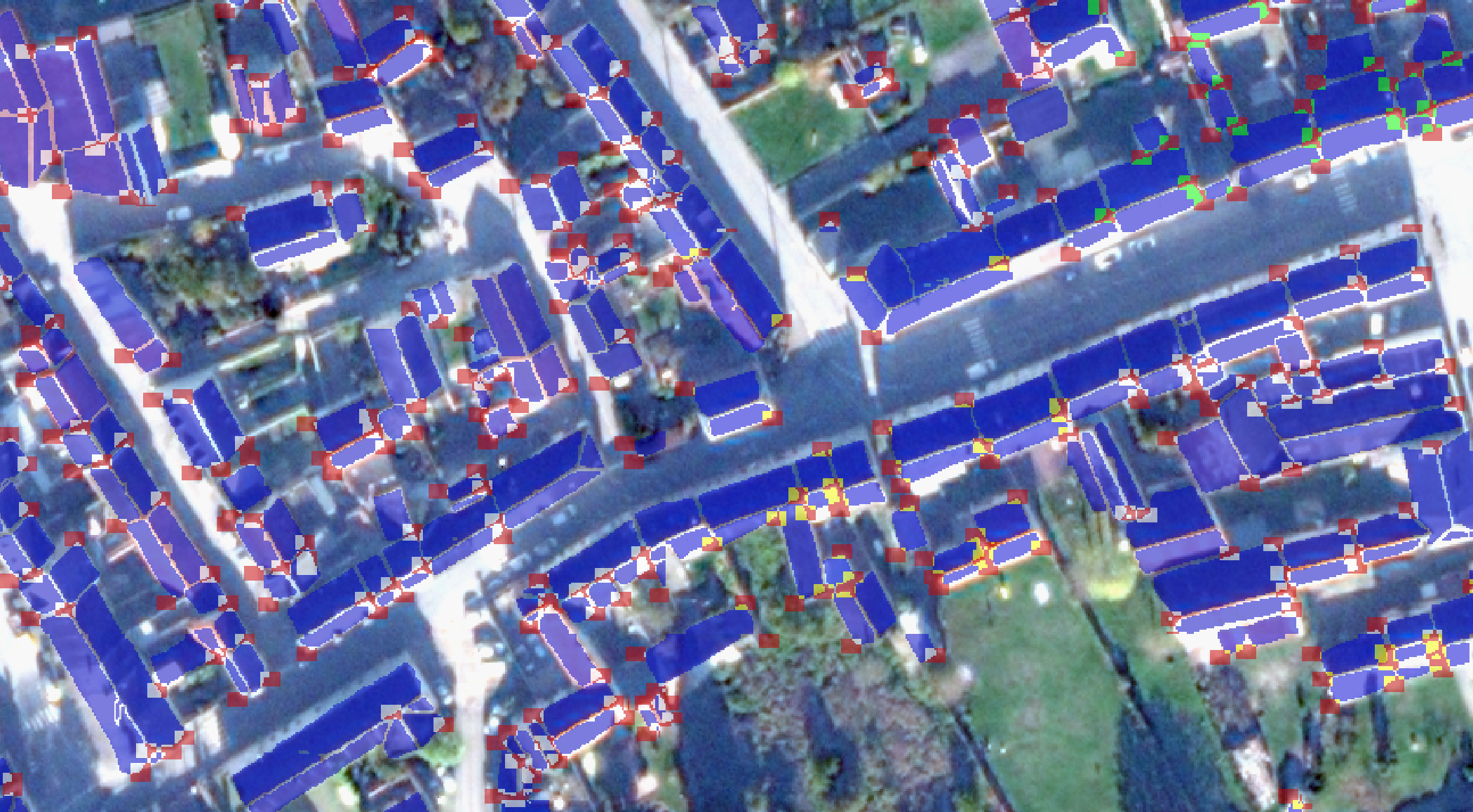}
\caption{\label{B2b} Outputs of the roof section keypoints inference before data postprocessing, overlaying the associated raster image of Sissonne. Each red square contains the height-to-ground $z$ of its associated roof corner within the red channel of the RGB picture, with a value $200+z$. For visualization purposes, when these overlap the blue segmented pixels of their associated roof section(s), they take a white, green, or yellow color if they belong to the training, validation, or testing set, respectively.}
\end{centering}
\end{figure}

The results of the 3D inference on all data sets (training, validation, testing) are shown in Fig. \ref{B2a} for Mourmelon-le-Grand and Fig. \ref{B2b} for Sissonne. The 3D inference results are represented via color-coded roof section corners, each color code being derived from a unique class corresponding to the discrete corner's height-to-ground in meters. This visual representation and panoptic segmentation allows us to qualitatively evaluate the model's ability to infer the 3D structure of the urban landscape from the 2D segmentation output. It is noteworthy that the model exhibits a high level of detail in the 3D inference, successfully capturing the complex architectural features and the varying heights of the buildings in both cities.

\begin{figure}[htbp]
\begin{centering}
\includegraphics[scale=0.26]{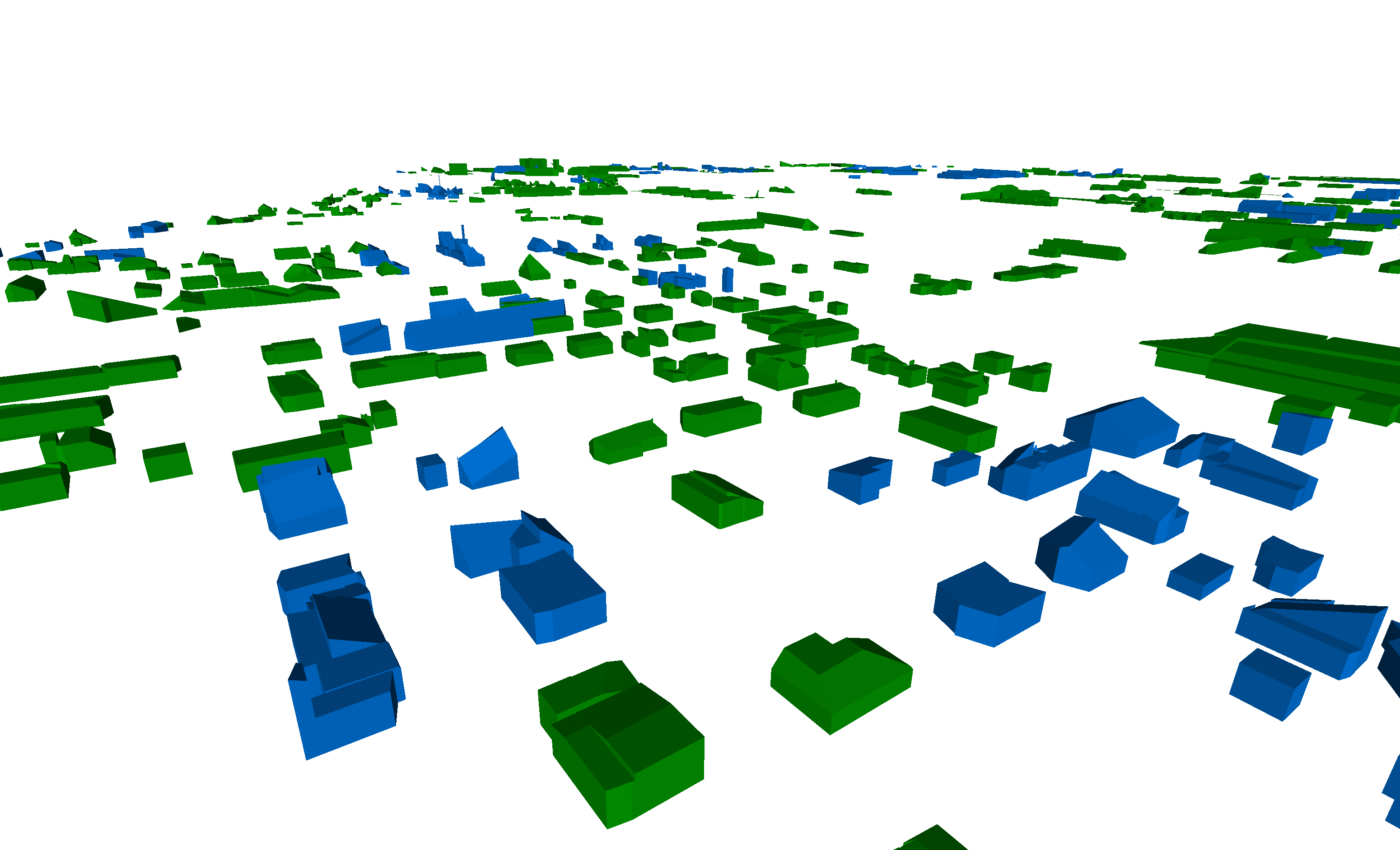}\\
\includegraphics[scale=0.26]{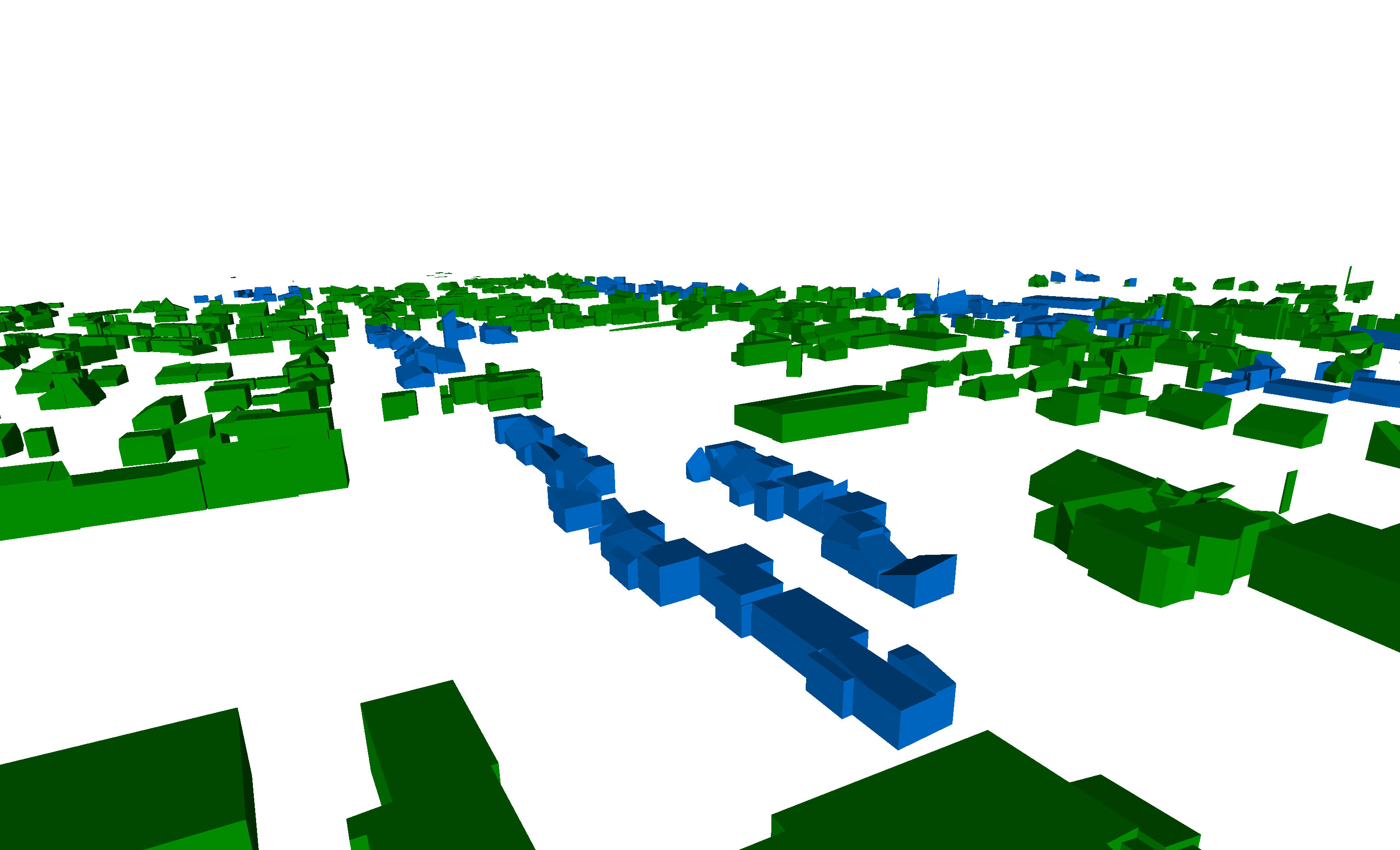}\\
\includegraphics[scale=0.26]{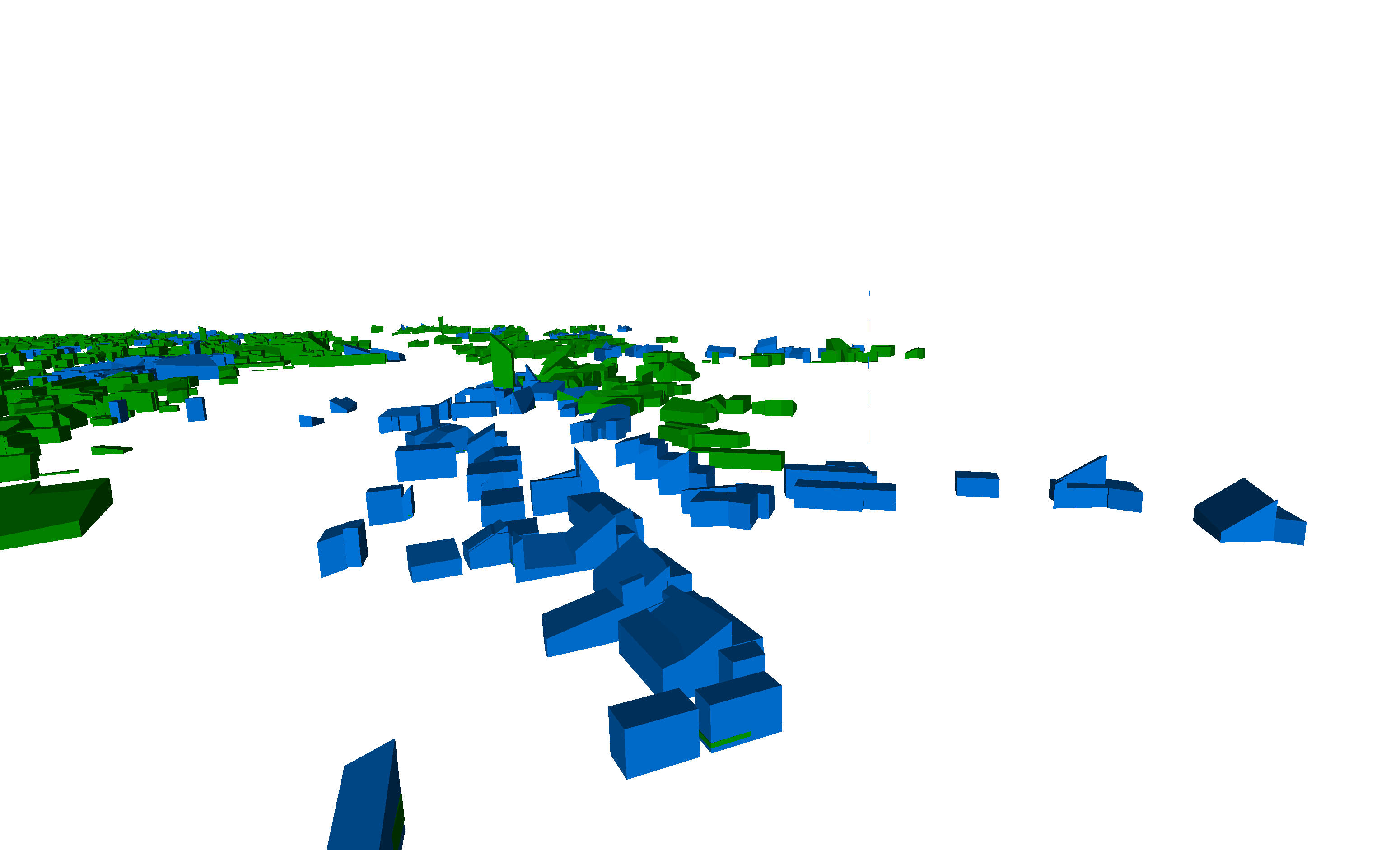}
\caption{\label{B3a} Visualization of the 3D reconstruction output of the testing set of Mourmelon-le-Grand with corrected DTM values. The training and validation data is displayed in green, and the KIBS model output in blue.}
\end{centering}
\end{figure}

\begin{figure}[htbp]
\begin{centering}
\includegraphics[scale=0.26]{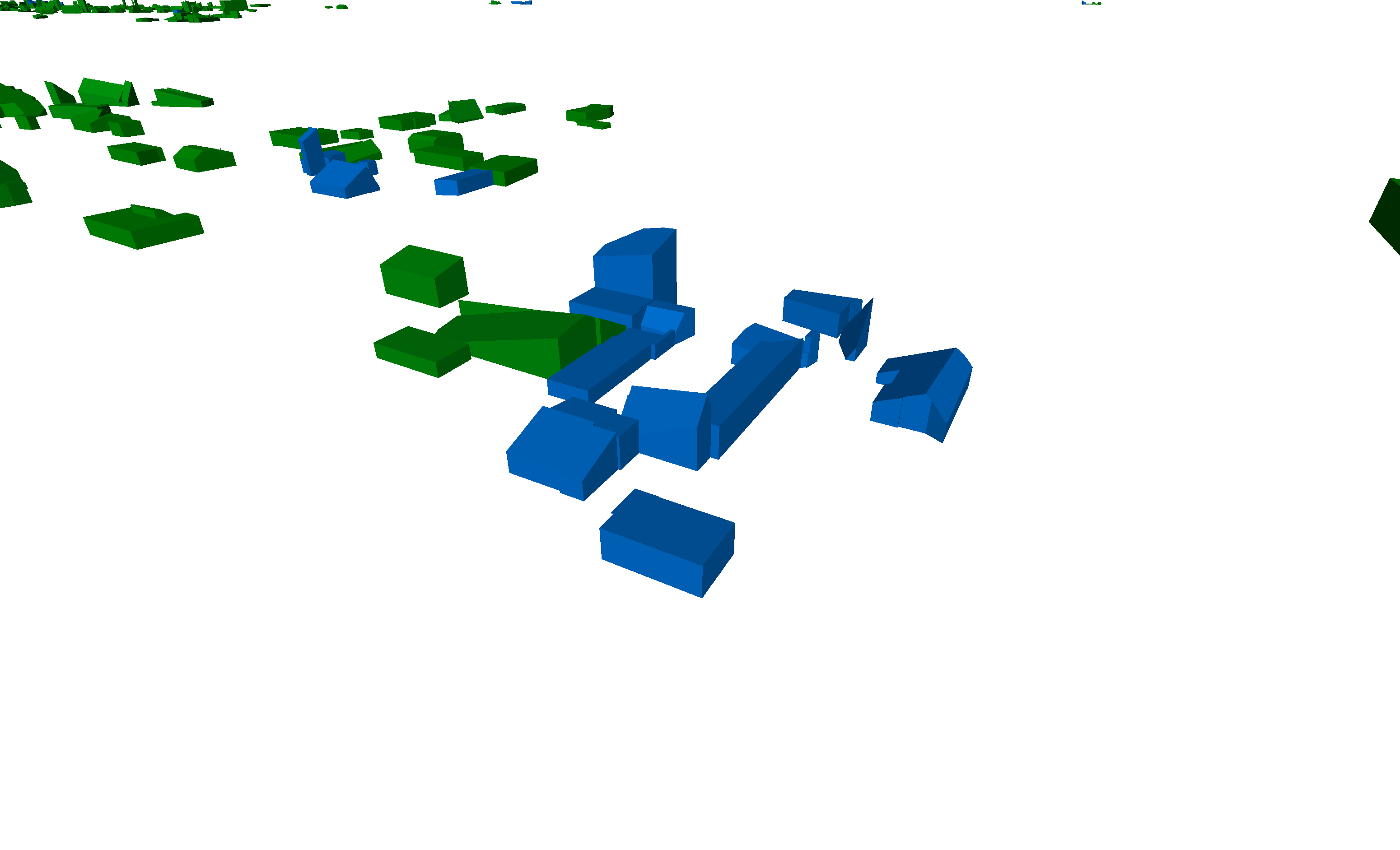}\\
\includegraphics[scale=0.26]{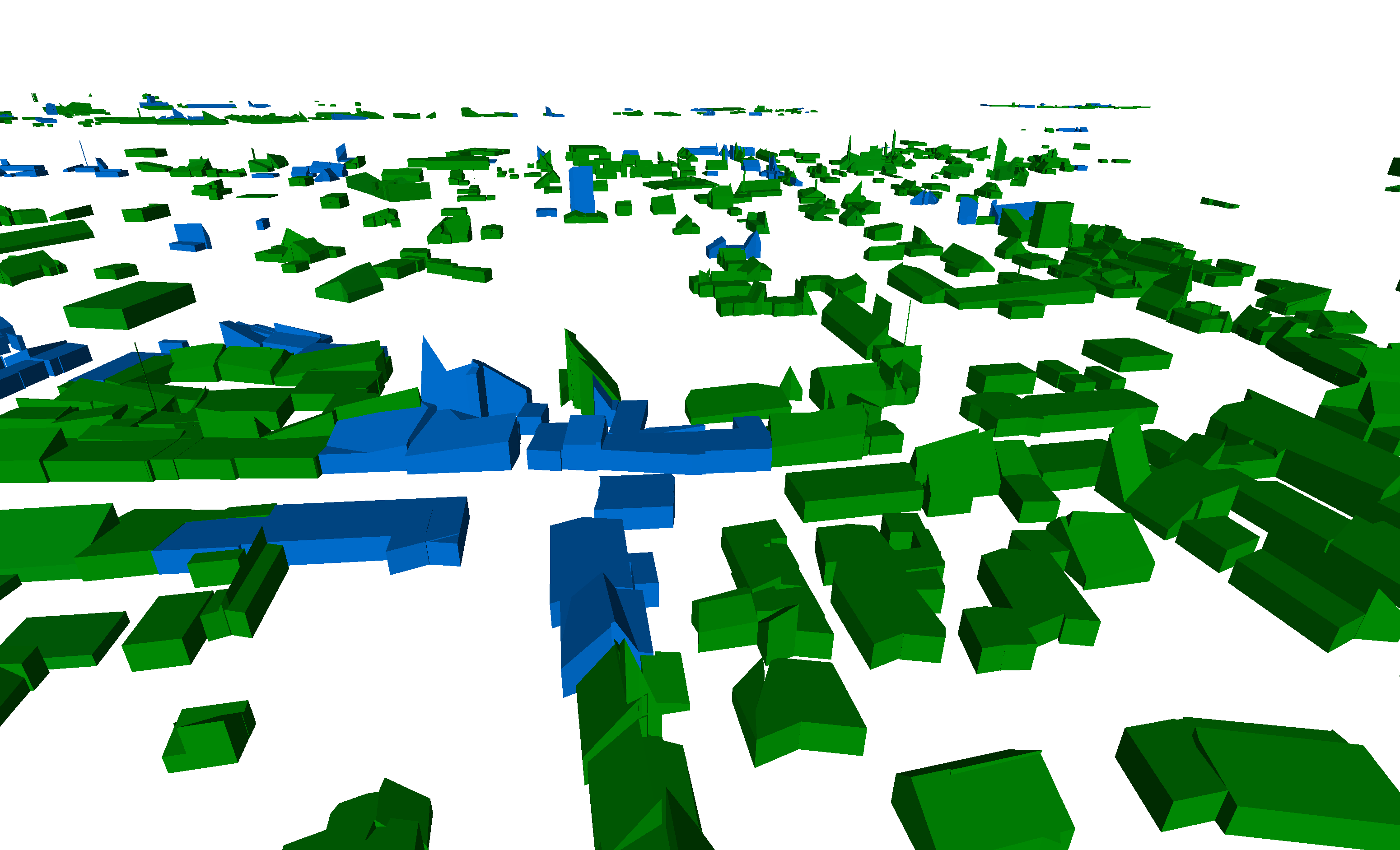}\\
\includegraphics[scale=0.26]{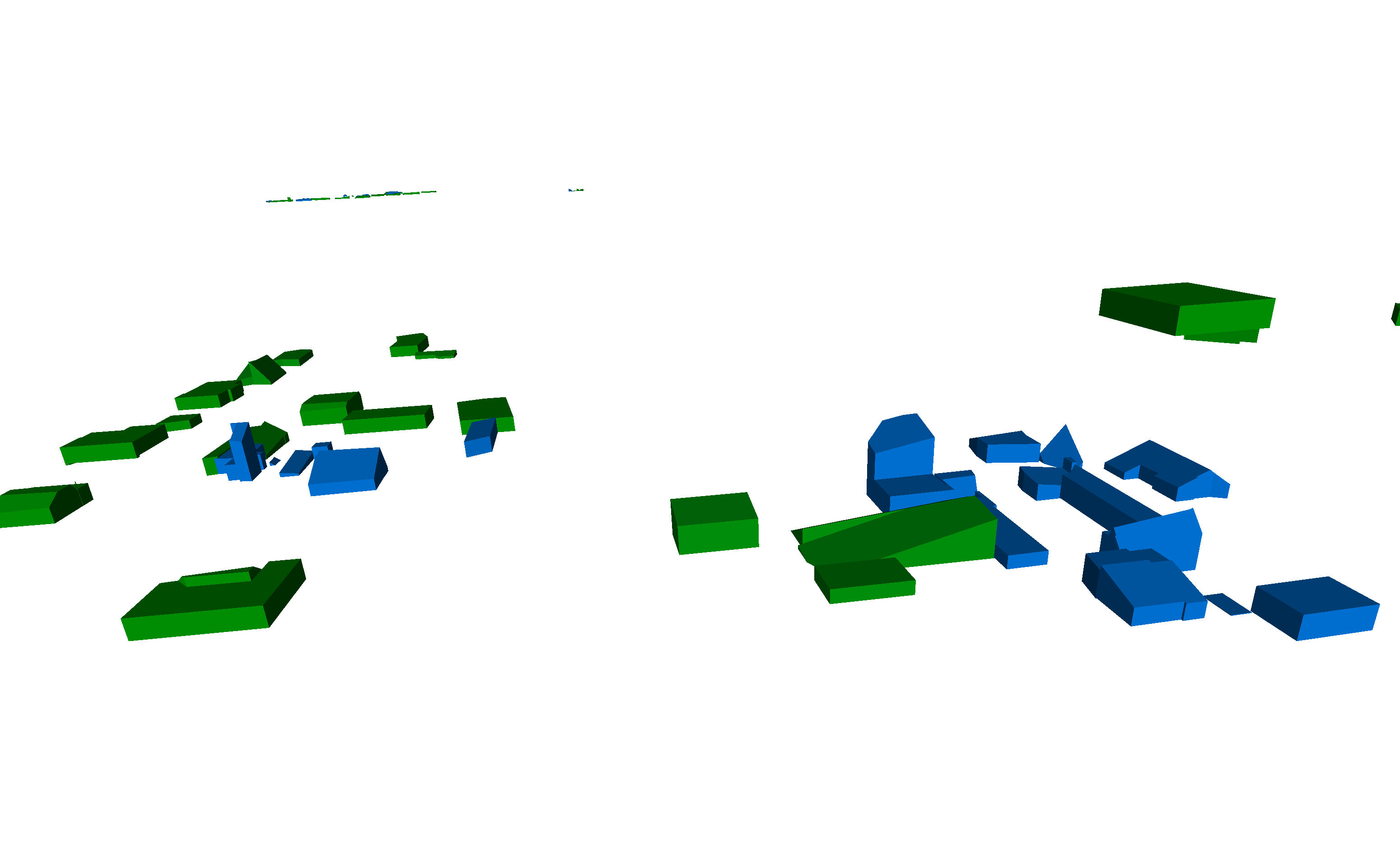}
\caption{\label{B3b} Visualization of the 3D reconstruction output of the testing set of Sissonne with corrected DTM values. The training and validation data is displayed in green, and the KIBS model output in blue.}
\end{centering}
\end{figure}

The visualization of this 3D inference, scaled to DSM values, is displayed after the KSR reconstruction~\cite{Yu2022} in Fig. \ref{B3a} for Mourmelon-le-Grand and \ref{B3b} for Sissonne. This provides a more tangible and intuitive understanding of the model's output, effectively transforming the aforementioned panoptic segmentation into a 3D model of the urban landscape, not only for the roof structures but for the whole buildings underneath. 

\subsection{Quantitative results} 
\label{SectionIVb}

The results of the KIBS model are shown in Tab. \ref{T1a} for Mourmelon-le-Grand and Tab. \ref{T1b} for Sissonne.  The results of the 2D segmentation can be summed up through the Jaccard index, also called Intersection over Union (IoU), which is the percentage of the $M$ accurately segmented pixels on the 2D map, with respect to the ground truth pixels. We obtain an IoU of $88.55 \%$ for the testing set. 

The accuracy of the 3D inference can be summed up \textit{for these pixels that were correctly 2D segmented wrt. ground truth}, through their heights mean accuracy, and mean square error. The heights mean accuracy is the average of the absolute differences between the heights of each correctly segmented pixels $\hat{z}$, and the height of its associated ground truth pixels $z$, expressed as a percentage of the latter: $\Sigma_{i=0}^M 100|\hat{z}_i - z|/zM$. We find a heights' mean accuracy for the testing set of $74.85 \%$ for Mourmelon-le-Grand, and $72.57 \%$ for Sissonne. And we find a heights' mean value for the testing set of $1.60$ m for Mourmelon-le-Grand, and $2.06$ m for Sissonne. 

And one can study the 3D reconstruction efficiency via the mean square error, knowing our data set has an average roof height of $6.36$ m for Mourmelon-le-Grand, and $7.53$ m for Sissonne. The heights' mean square error is given by the average squared difference between the heights of each correctly segmented pixels $\hat{z}$, and the height of its associated ground truth pixels $z$: $\Sigma_{i=0}^M (\hat{z}_i - z)^2/M$. We thus find a heights' mean square error for the testing set of $2.35$ m$^2$ for Mourmelon, and $7.41$ m$^2$ for Sissonne. 

We can see from these two latter statistics, that the aim of urban 3D reconstruction at LOD2 is reached.

\begin{table}[htbp]
\centering
\begin{tabular}{ | m{4cm} | m{2cm}| m{2cm} | m{2cm} | } 
\hline
\textbf{Statistic (Mourmelon)} & \textbf{Training} & \textbf{Validation} & \textbf{Testing} \\ 
\hline
Jaccard index (IoU) & 88.56 $\%$ & 86.96 $\%$  & 88.55 $\%$ \\ 
\hline
Heights' mean accuracy & 76.87 $\%$ & 74.03 $\%$ & 74.85 $\%$ \\ 
\hline
Heights' mean difference & 1.47 m & 1.65 m & 1.60 m \\ 
\hline
Heights' mean square error & 1.99 m$^2$ & 2.51 m$^2$ & 2.35 m$^2$ \\  
\hline
\end{tabular}
\caption{\label{T1a} Results of the KIBS model for Mourmelon-le-Grand. The results of the 2D segmentation shown by the Jaccard index (i.e. Intersection over Union, or IoU), which is the percentage of the $M$ correctly segmented pixels compared to ground truth. For $\hat{z}$ and $z$ the heights-to-ground of these correctly segmented pixels and of ground truth pixels respectively, the results of the 3D reconstruction is shown by the heights' mean accuracy $\Sigma_{i=0}^M 100|\hat{z}_i - z|/zM$ and mean square error $\Sigma_{i=0}^M (\hat{z}_i - z)^2/M$.}
\end{table}

\begin{table}[htbp]
\centering
\begin{tabular}{ | m{4cm} | m{2cm}| m{2cm} | m{2cm} | } 
\hline
\textbf{Statistic (Sissonne)} & \textbf{Training} & \textbf{Validation} & \textbf{Testing} \\ 
\hline
Jaccard index (IoU) & 88.67 $\%$ & 73.49 $\%$ & 75.21 $\%$ \\ 
\hline
Heights' mean accuracy & 74.28 $\%$ & 71.08 $\%$ & 72.57 $\%$ \\ 
\hline
Heights' mean difference & 1.94 m & 2.18 m & 2.06 m \\  
\hline
Heights' mean square error & 7.21 m$^2$ & 8.29 m$^2$ & 7.41 m$^2$ \\  
\hline
\end{tabular}
\caption{\label{T1b} Results of the KIBS model for Sissonne. The results of the 2D segmentation shown by the Jaccard index (i.e. Intersection over Union, or IoU), which is the percentage of the $M$ correctly segmented pixels compared to ground truth. For $\hat{z}$ and $z$ the heights-to-ground of these correctly segmented pixels and of ground truth pixels respectively, the results of the 3D reconstruction is shown by the heights' mean accuracy $\Sigma_{i=0}^M 100|\hat{z}_i - z|/zM$ and mean square error $\Sigma_{i=0}^M (\hat{z}_i - z)^2/M$.}
\end{table}

\subsection{Performance} 
\label{SectionIVc}
Let $s \in \mathbb N^{\ast}$ be the number of pixels giving the (squared) raster tile images' size (e.g. for $s=230$, the raster tile images are of size $s \times s = 230 \times 230$), $p \in \mathbb N^{\ast}$ the number of pixels of the raster tile images' margin overlap, and $q \in \mathbb N^{\ast}$ the number of pixels of the size of the segmented roof section corners (e.g. for $q=15$, the roof section corners were segmented as red squares of sizes $q \times q = 15 \times 15$). The KIBS model performance has been explored through several combinations of the model hyperparameters on the validation set, by visualizing the output results for combinations of these hyperparameters $s, p, q$. 

\begin{figure}[!htbp]
\begin{centering}
\includegraphics[scale=0.123]{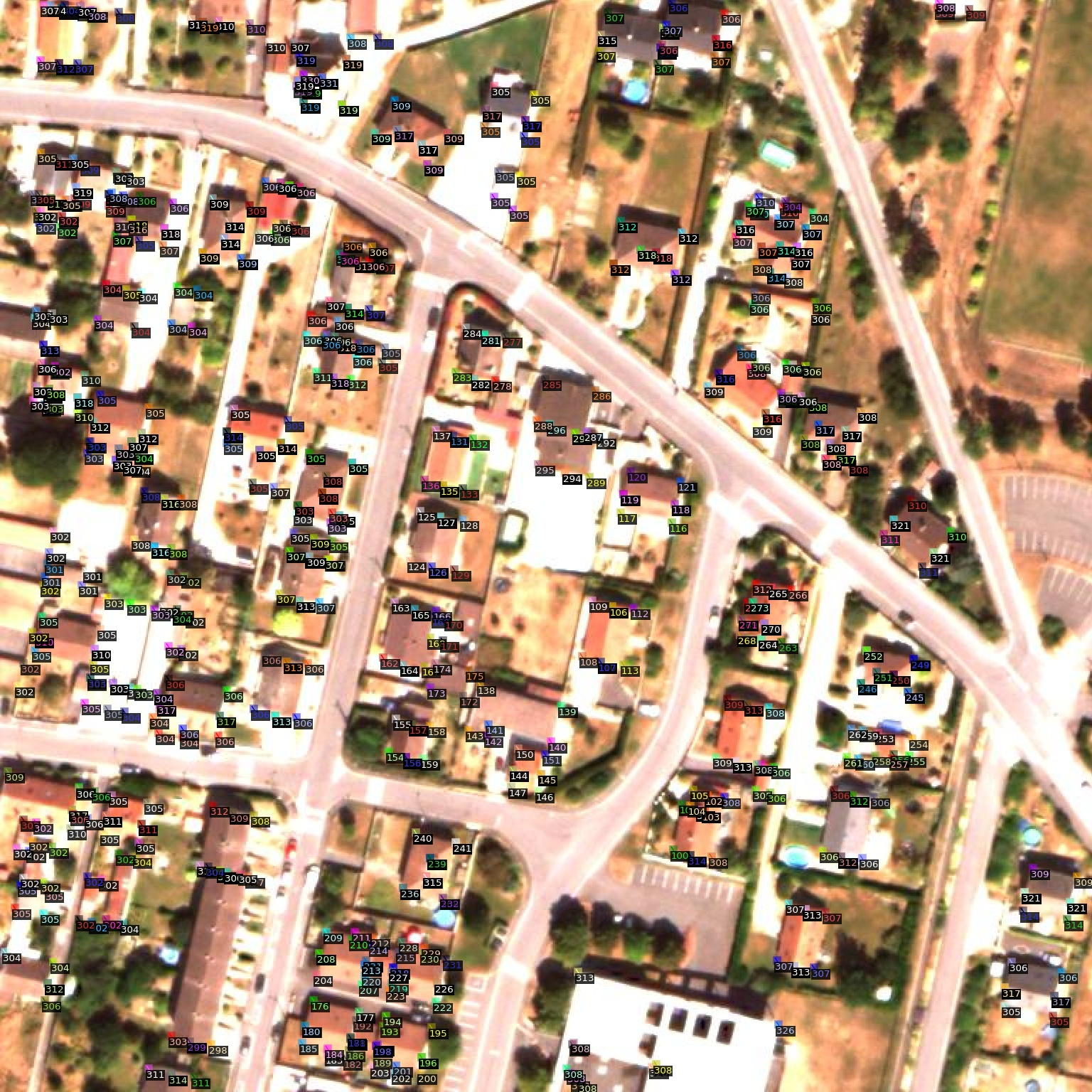}
\includegraphics[scale=0.34]{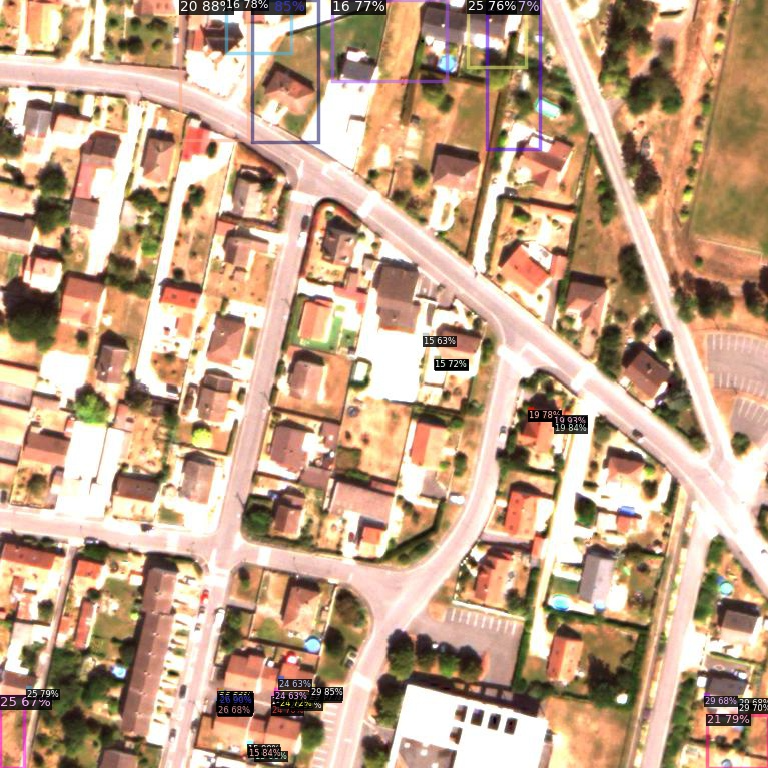}
\caption{\label{S0} Example of roof section corners inference directly on a raster image of size $768 \times 768$, with ground truth on the left, and result output on the right. The large number of roof sections in the ground truth overloads the Mask-RCNN algorithm.}
\end{centering}
\end{figure}

The change in performance for different raster tile images' sizes $s$ was explored, with values $s = \{150, 230, 300, 768\}$. We found the use of larger resolution raster images as input to be a limiting factor to the number of roof section corners that could be detected by the second Mask-RCNN model (as shown in the Supplementary Material section \ref{SectionVIe} by comparison with Fig. \ref{S0}, where $s=768$, or on Fig. \ref{S1}, where $s=300$). We thus found a better performance for smaller resolutions, especially at $s = 230$ (calculated for values $p=10$ and $q=15$ only). 

Secondly, another explored hyperparameter was the raster tile images' margin overlap $p$, with values $p = \{10, 150\}$ pixels (calculated for values $s=230$ and $q=15$ only). We found a large margin overlap value to cause intractable memory issues during the run time, and hence selected $p = 10$. 

Thirdly, the size of the segmented roof section corners $q$ was explored, with values $q= \{10, 15\}$ (for values $s=230$ and $p=10$ only). As said, this hyperparameter has a great impact on the overall KIBS model performance, since too large squares may assign the height of a given corner to several others as well, and too small squares may produce false negatives by not overlapping their associated segmented roof sections at postprocessing, We thus found better performance for $q = 15$ (calculated for values $s=230$ and $p=10$ only).

\subsection{Limitations} 
\label{SectionIVc2}

As said, the core premise behind the KIBS model hypothesis is that the oblique perspective of the satellite raster image supplies the deep learning algorithms with valuable and complex information related to the roofs corners height-to-ground. This includes aspects such as the tilt of the buildings' walls, the shadows cast by the buildings, the perspective of the roof peak or ridge, and so on. These elements collectively enable the algorithms to deduce the height-to-ground of the corners of the roof sections with a level of accuracy that meets the standards of the LOD2 requirement. This is an important feature of the KIBS prior pertaining to its generalization, because each data set used to train the 3D reconstruction part of the model has its own specific buildings inclinations (related to the raster' satellite viewing angle $\alpha$), and its own specific shading of the buildings (related to the raster' solar zenith angle $\theta$), as aforementioned for a our satellite data sets. The KIBS method trained on a data set with such angles $\alpha$ and $\theta$, should hence only generalize to new raster sets taken with angle parameters lying in the neighborhoods of those of the training set, so that the variations in the model inference of the buildings' height-to-ground are negligible within the requirements of a LOD2 precision range.

\subsection{Baseline comparison} 
\label{SectionIVd}

Due to the uniqueness of the results of this study, the KIBS method faces a challenge in finding relevant methods for a useful baseline comparison. Other interesting research works like~\cite{Leotta2019} (which does 3D urban reconstruction at LOD1), and~\cite{Ren2021,patentluxcarta} (which both use roof primes for the urban reconstruction) rely on third parties code which is not accessible. But a rigorous approach can be to use the 2D segmentation step of our KIBS approach, and then assign the segmented pixels' height-to-ground via another DSM, courtesy of LuxCarta, which is of LOD1. The resulting point cloud can then be approximated as roof sections unto 3D reconstruction, as shown on Fig. \ref{S4} below, with a rather poor precision. 

\begin{figure}[htbp]
\begin{centering}
\includegraphics[scale=0.25]{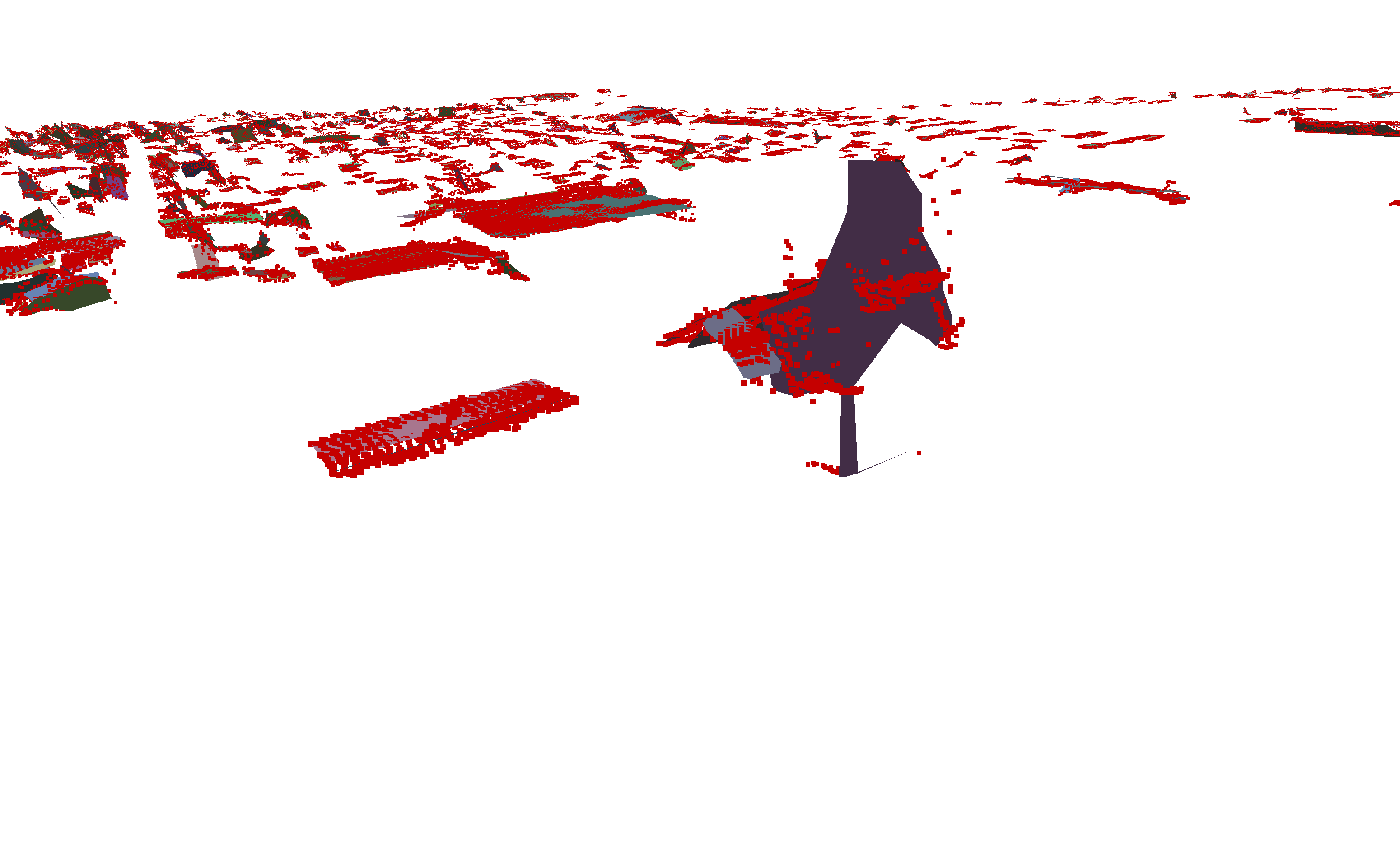} \
\includegraphics[scale=0.25]{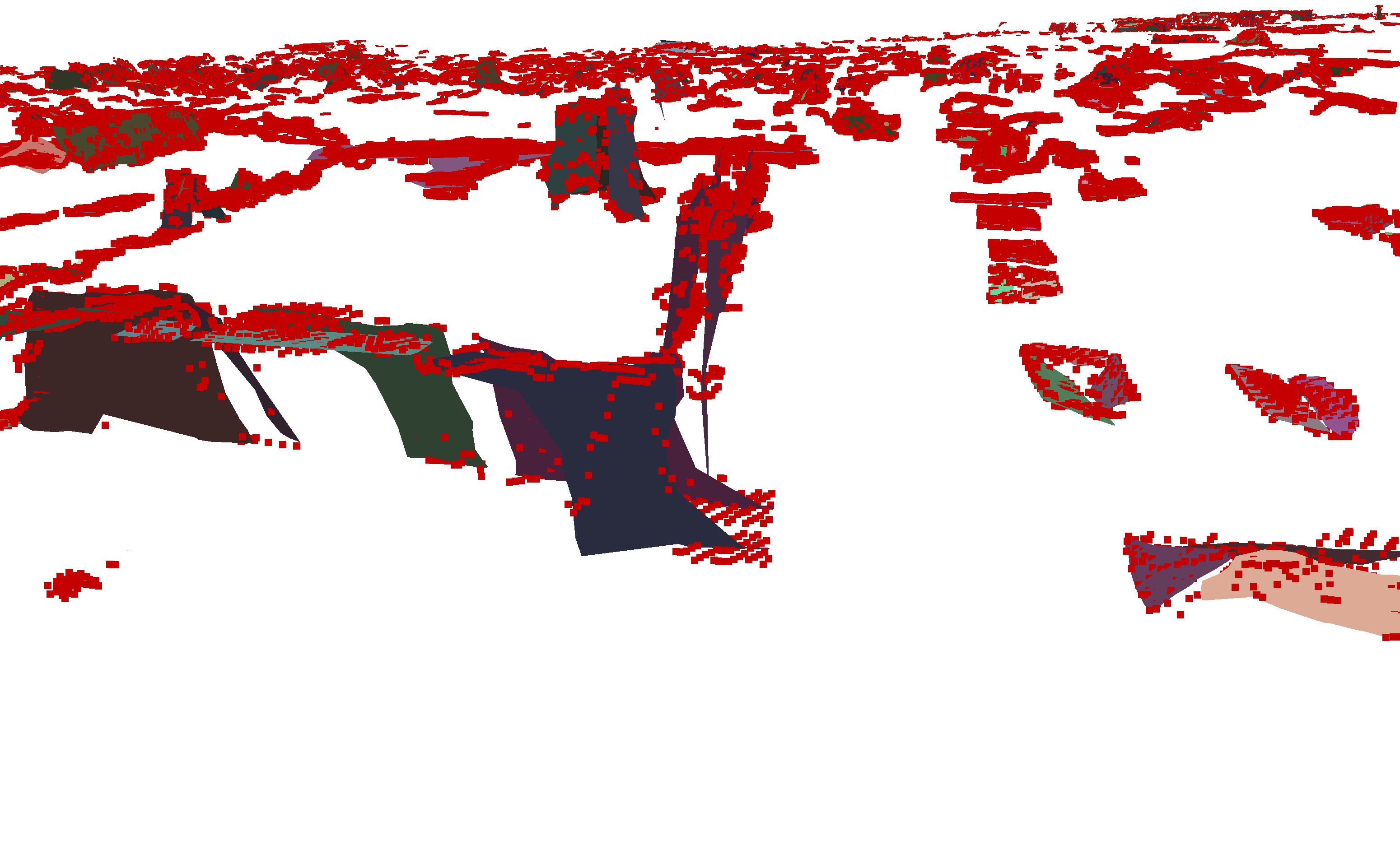} \
\includegraphics[scale=0.25]{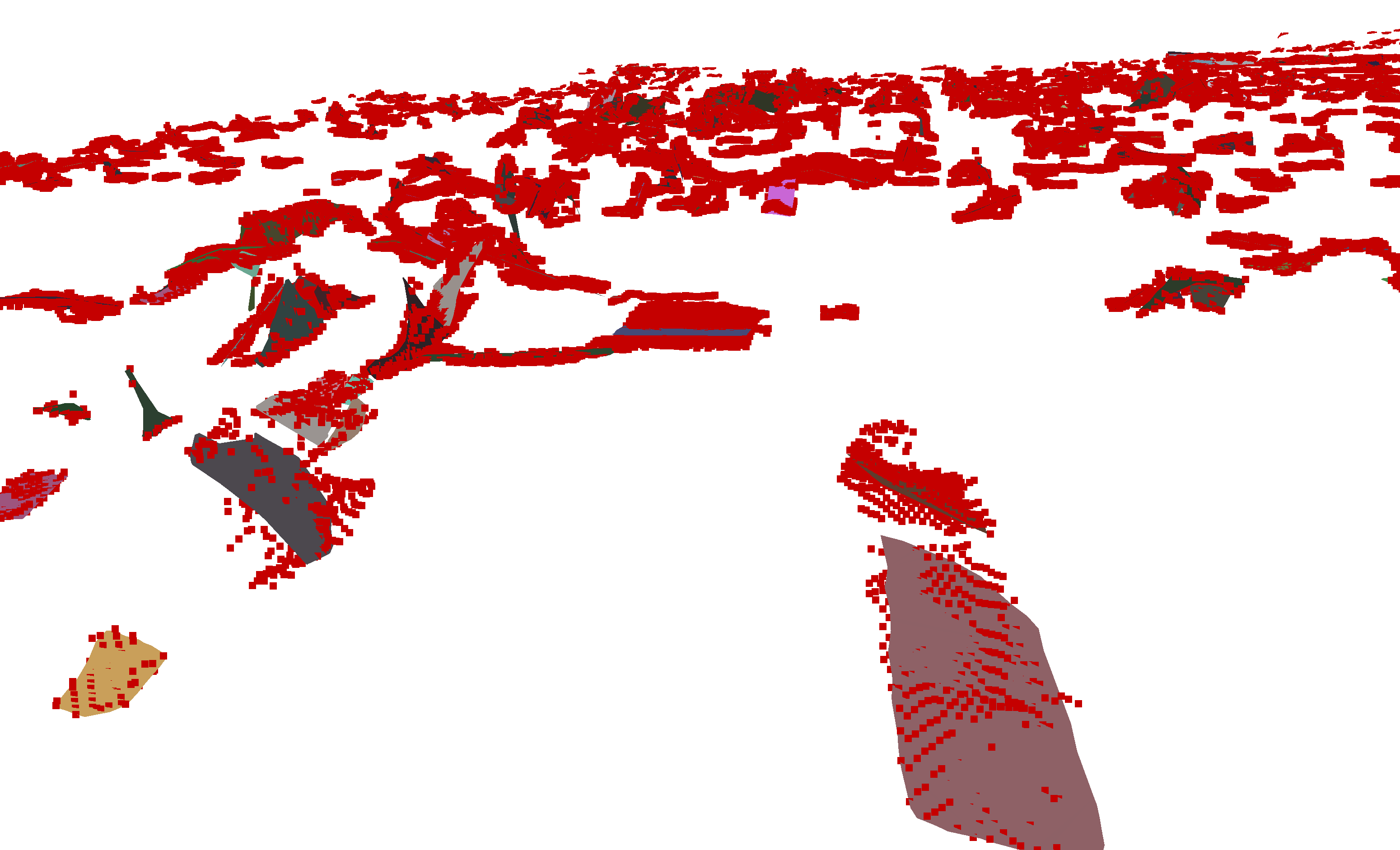}
\caption{\label{S4} Comparison baseline based on the 2D segmentation of our KIBS method, followed by the interpolation of the elevation point cloud stemming from our DSM data (red points), courtesy of LuxCarta, with ensuing roof section 3D reconstruction (coloured planes).} 
\end{centering}
\end{figure}

\subsection{Ablation study} 
\label{SectionIVe}

We have used different ablations of the model and studied its change in performance. 

Firstly, if one tries to infer the roof section corners' position and heights $(x, y, z)$ directly through the raster satellite image (cf. images on the left of Fig. \ref{S1} of section \ref{SectionVIe} in the Supplementary Material), the result output shows a poor performance. 

Secondly, when the blending of the 2D segmentation output is performed on either the red or green channels of the RGB raster image, the results and model accuracy do not change much with our current approach (cf. images in the center of Fig. \ref{S1} of section \ref{SectionVIe} in the Supplementary Material). 

Thirdly, if one tries to improve the 2D location $(x, y)$ of the roof section corners by an image input consisting in the direct raster satellite image, or in the binary masks of the roof sections (cf. images on the right of Fig. \ref{S1} of section \ref{SectionVIe} in the Supplementary Material), the output results are very unsatisfying. 

Fourthly, if one tries and infers the roof section corners' heights out of a blending of three stereo satellite pictures of the same geographical area, each taken with different satellite viewing angle and solar zenith angle, the output results show the poor performance of this approach (cf. Fig. \ref{S2} of section \ref{SectionVIe} in the Supplementary Material). 

Fifthly, other very different models and algorithms than the Mask-RCNN solution were tried and used in this research, both for the 2D segmentation part and the 3D reconstruction part with the work of~\cite{Yu2019,Lenssen2020}, but as can be shown on Fig. \ref{S3} of section \ref{SectionVIe} in the Supplementary Material, this approach failed completely. 

Sixthly, likewise, a RegNet architecture~\cite{Xu2022} of the Detectron2 model zoo (\verb?regnety_4gf_dds_FPN?) has been used instead of the Mask-RCNN blocks, but with poor results in our time-constrained hyperparameter space optimization procedure so far.

\section{Conclusion}
\label{SectionV}

We have thus presented a new method named KIBS for the urban 3D reconstruction of satellite images at a LOD2, with two central features: an end-to-end deep learning approach, and a model input based on a one-shot satellite raster image. The backbone of this deep learning model is a two-step method relying firstly on a Mask-RCNN algorithm performing the 2D segmentation of the individual roof sections, and secondly on another Mask-RCNN algorithm of exact same architecture using the latter output blended into the raster image in order to infer the roof section corners and their heights. 
The performance of this KIBS approach is displayed by a Jaccard index for the 2D segmentation of the roof sections of $88.55 \%$ (Mourmelon-le-Grand) and $75.21\%$ (Sissonne), and a heights' mean value for the roof section pixels correctly inferred by the 2D segmentation method of $1.60$ m (Mourmelon-le-Grand) and $2.06$ m (Sissonne). The KIBS method can thus perform 3D reconstruction of urban satellite raster images within the requirements of the LOD2 precision range. 

As such, the authors posit that the weight played by deep learning methods in satellite and aerial data ground reconstruction, whether via end-to-end approaches or in complement of more procedural approaches, will only increase in coming years. 

Bearing in mind the time-constrained optimization procedure of the method presented in this research work, the authors also posit that the performance results of the KIBS method may be easily enhanced at a little cost, notably by a further exploration of the hyperparameter space, and by use of deep learning architectures other than the Mask-RCNN neural networks here employed. 

Especially, a direct natural extension of the KIBS approach should study whether one single neural network comprising this two-step approach (2D segmentation followed by 3D reconstruction) into one backbone architecture could be designed. The authors posit this general monocular or single-shot approach to 3D inference could find many promising applications reaching far beyond the satellite and aerial imagery segments of computer vision, and pertain to all 3D inference methods of machine learning in the largest sense, with other potential applications in autonomous driving, drones engineering, environmental monitoring, and virtual reality. 

This said, our work also raises new questions, such as how to further improve the accuracy of 3D inference, how to handle taller structures, and how to apply our methods to other types of data. A crucial future prospect of the KIBS method pertains to its generalization, not only for very different data sets (e.g. dense city centers with tall buildings), but also wrt. to raster data sets of different satellite viewing angle $\alpha$ and solar zenith angle $\omega$ from those of our training set. Thanks to its short computational training and inference times, a suite of several KIBS algorithms could be trained on sets of data taken with different combinations of these two angle values' neighborhoods, so as to reach a practical generalisation threshold by modular learning, corresponding to the offers of satellite data vendors.

\clearpage

\section{Supplementary material}
\label{SectionVI}

\subsection{Implementation details of the 2D segmentation part}
\label{SectionVIa}

The training data preprocessing for the Mask-RCNN model performing the 2D segmentation of the roof lines first relies on first slicing the overall $8687 \times 9890$ satellite raster image into individual tiles of $230 \times 230$ individual raster images. These are cut to overlap each other on all four sides by a margin of $10$ pixels, to improve the future reconstruction at inference level. 

Then, the ground truth shapefile of the polygons delimiting each roof section contours is extracted for each associated $230 \times 230$ tile raster image. For each such tile raster image, a set of $230 \times 230$ black and white images is generated for each roof section, where each white pixel belongs to one unique roof section per image, and all other pixels are set as black. Each roof section mask is given one same dummy class label at this stage. 

All these generated black and white images associated with each tile raster image are given a unique file name that allows a specific \verb?pycococreator? algorithm~\cite{pycococreatorgithub} to generate a .json file for these ground truth masks in the PYCOCO format~\cite{pycoco2014}. 

The set of all such tile raster images, together with their associated ground truth images is then shuffled randomly according to a uniform distribution in order to build three disjoints sets: one for training ($60\%$ of the whole data set), one for validation ($20\%$ of the whole data set), and one for testing ($20\%$ of the whole data set). 

Via \verb?pycococreator?, a .json ground truth file associated with the training set is hence generated, and likewise for the validation and testing sets. 

All the tile raster images and these generated .json ground truth files are then given as input to train a Mask-RCNN artificial neural network~\cite{He2017} from the Detectron2 suite~\cite{detectron2github} named \verb?mask_rcnn_R_50_FPN_3x?. This network consists in a backbone combination of a ResNet-50 model~\cite{He2016} stacked with a Feature Pyramid Network~\cite{Lin2017} (FPN), comprising standard convolution and fully-connected heads for mask and box prediction, respectively. It is pretrained with a $3$x schedule, corresponding to about $37$ COCO epochs. 

The results, presented in Section \ref{SectionIV}, are based on a six days training, on a Dell T630 GPU node of dual-Xeon E5-26xx with four GeForce GTX 1080 Ti GPUs cards, 3584 CUDA cores per card, and 11 GB of RAM capacity per card. 

The training metrics are shown in Fig. \ref{A2a}-\ref{A5b} of Section \ref{SectionIVc} below, and were monitored online via TensorBoard~\cite{tensorflow2015} in order to limit regularization issues. The weights of the Mask-RCNN network trained for this 2D segmentation are available on the KIBS GitHub repository~\cite{biyologithub}.

\subsection{Implementation details of the 3D reconstruction part}
\label{SectionVIb}

From a deep learning perspective, the 3D reconstruction training relies on a Mask-RCNN model of exact same architecture as for the 2D segmentation, but designed for panoptic segmentation, i.e. both pixel segmentation and class inference. In our case, the pixel segmentation here consists in the model drawing a $15 \times 15$ pixels square over each roof section corner of a raster image blended with the output of the 2D segmentation, and the class inference consists in giving each such corner a unique class label allowing to retrieve the corner's height-to-ground in meters. 

As said, after training, the output of the latter 2D segmentation is blended within the original associated RGB raster image, such that each segmented pixel (identifying a roof section) is given a value $\{0, 0, 200 \}$ if belonging to an image from the training data set, $\{0, 0, 210 \}$ if belonging to an image from the validation data set, and $\{0, 0, 220 \}$ if belonging to an image from the testing data set. Roof sections on the raster images hence appear in blue color. Ablation studies (see below) show this method allows the 3D reconstruction algorithm to identify much more efficiently the roof sections' corners, than if the ground truth was associated with the original raster images only. 

In our code (for particular reasons related to the Detectron2 framework), class labels are: \verb?hah? for a height of $1$ m, \verb?hbh? for $2$ m, \verb?hch? for $3$ m, $\dots$, \verb?hsh? for $19$ m. This range of $19$ possible different classes is due to the maximum corner's height in our particular data set not exceeding $19$ m above ground, but one can extend the number of these classes/heights much more in the Detectron2 framework to cope with the potential taller building structures of other data sets. 

Hence, if our data set contained skyscrapers or buildings of greater heights, the KIBS method and training could remain similar by simply increasing the number of possible classes, and/or raising the height granularity above $1$ m, and/or using non-linear graduations in the heights increments, etc. 

This said, the training, validation, and testing sets generation is done in a similar way as for the 2D segmentation: each aforementioned $230 \times 230$ blended raster image with a margin overlap of $10$ pixels on all four sides, is associated with a set of ground truth images, each of them representing on a black background a square of $15 \times 15$ white pixels, in order to represent a roof corner on this image. Its class name (i.e. corner's height) is given in the image file name. 

This is likewise proper and fed to the pycococreator framework, in order to produce .json annotation files for this ground truth data in a PYCOCO format that is understandable to the Detectron2 framework. A same Mask-RCNN model (\verb?mask_rcnn_R_50_FPN_3x?) as before is thus trained on this training data set, so as to identify roof corners and their classes (i.e. heights). The results, presented in Section \ref{SectionIV}, are based on a six days training, also on the same hardware as before (four GeForce GTX 1080 Ti GPUs cards). 

The training metrics are shown in Fig. \ref{A2a}-\ref{A5b} of Section \ref{SectionIVd} below, and can be monitored online in order to limit regularization issues. The learning weights of the Mask-RCNN model for this 3D reconstruction are available on the KIBS GitHub repository~\cite{biyologithub}.

\subsection{Training and validation metrics}
\label{SectionVIc}

\begin{figure}[!htbp]
\begin{centering}
\includegraphics[scale=0.2]{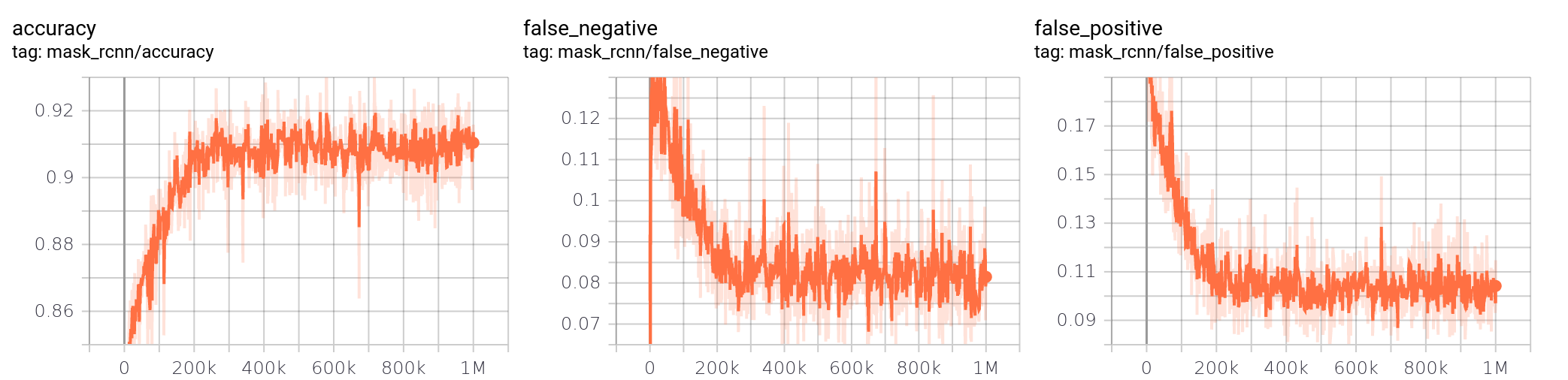}
\caption{\label{A2a} Training of the 2D segmentation with Mask-RCNN for Mourmelon-le-Grand, showing the model's precision (left), error rates for false negatives (center) and false positives (right), as a function of the number of epochs (x-axis). These plots were monitored online via TensorBoard~\cite{tensorflow2015} in order to limit regularization issues.}
\end{centering}
\end{figure}

\begin{figure}[!htbp]
\begin{centering}
\includegraphics[scale=0.2]{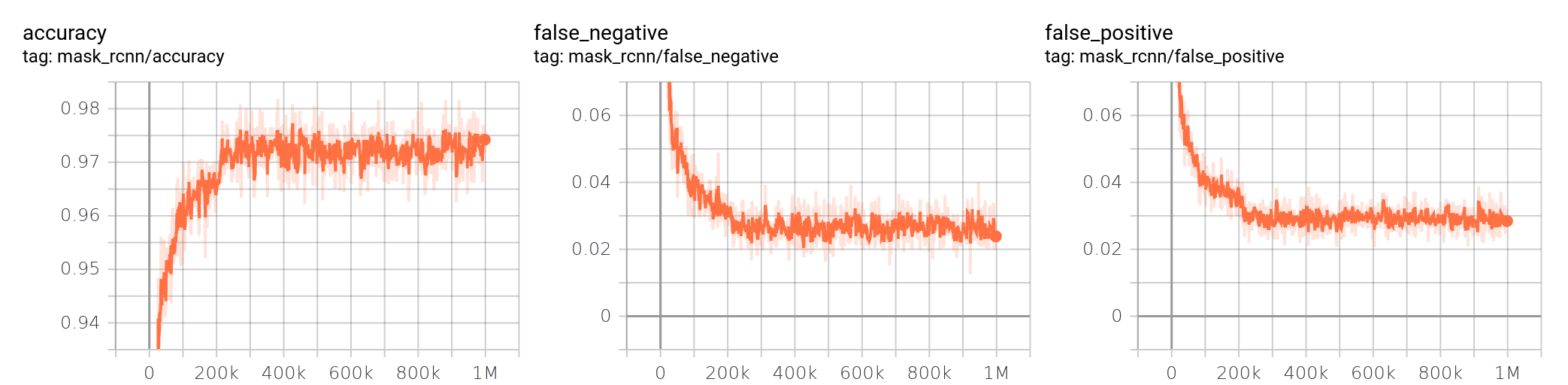}
\caption{\label{A2b} Training of the 2D segmentation with Mask-RCNN for Sissonne, showing the model's precision (left), error rates for false negatives (center) and false positives (right), as a function of the number of epochs (x-axis). These plots were monitored online via TensorBoard~\cite{tensorflow2015} in order to limit regularization issues.}
\end{centering}
\end{figure} 

\begin{figure}[!htbp]
\begin{centering}
\includegraphics[scale=0.2]{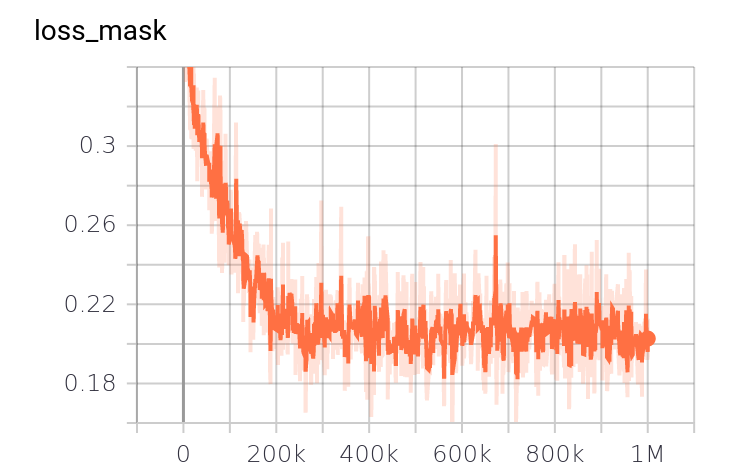}
\includegraphics[scale=0.2]{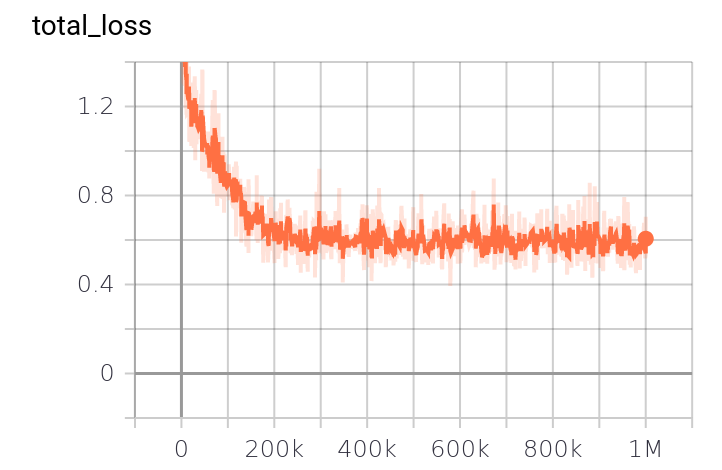}
\caption{\label{A3a} Training of the 2D segmentation with Mask-RCNN for Mourmelon-le-Grand, showing the model's mask loss (left) and total loss (right), as a function of the number of epochs (x-axis). These plots were monitored online via TensorBoard~\cite{tensorflow2015} in order to limit regularization issues.} 
\end{centering}
\end{figure} 

\begin{figure}[!htbp]
\begin{centering}
\includegraphics[scale=0.2]{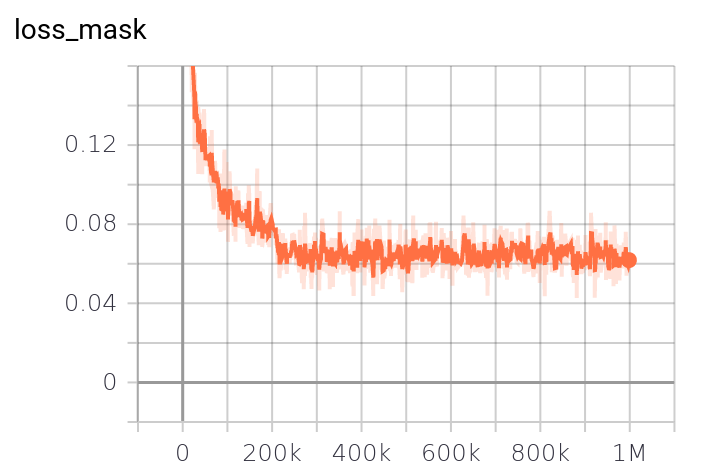}
\includegraphics[scale=0.2]{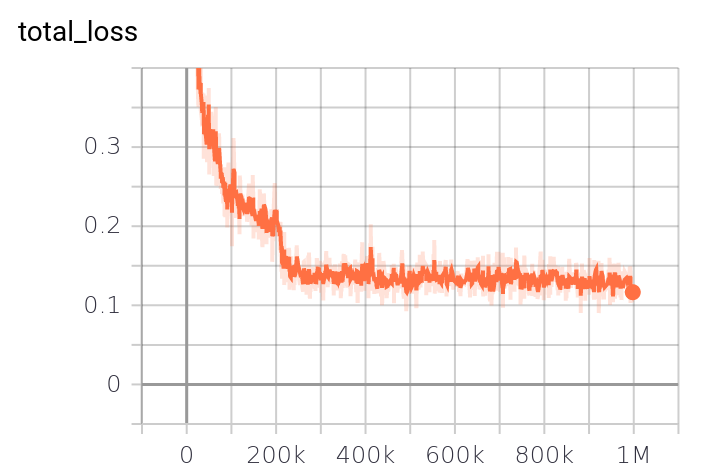}
\caption{\label{A3b} Training of the 2D segmentation with Mask-RCNN for Sissonne, showing the model's mask loss (left) and total loss (right), as a function of the number of epochs (x-axis). These plots were monitored online via TensorBoard~\cite{tensorflow2015} in order to limit regularization issues.} 
\end{centering}
\end{figure} 

\begin{figure}[!htbp]
\begin{centering}
\includegraphics[scale=0.2]{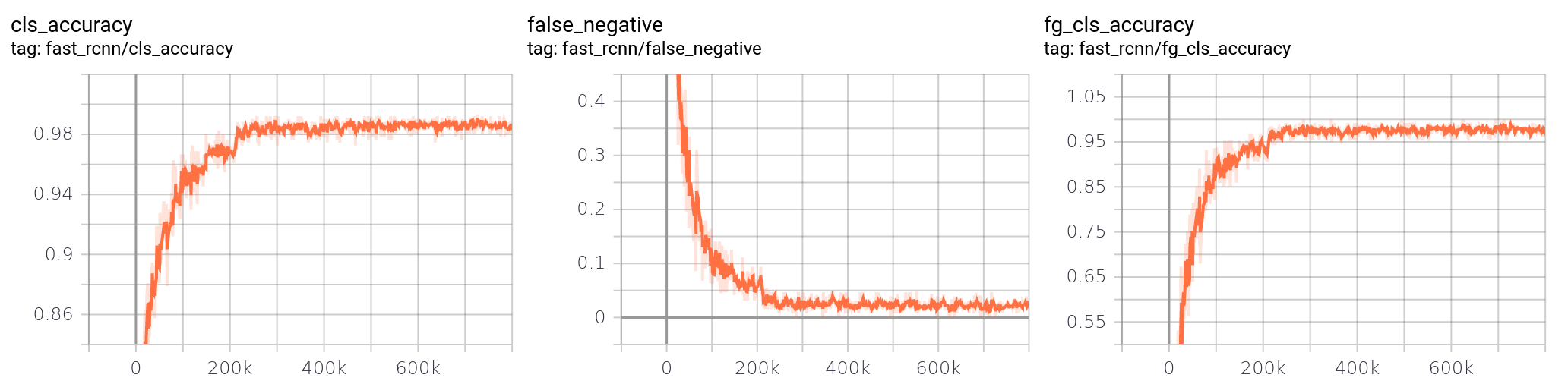}
\caption{\label{A4a} Training of the 3D reconstruction with Mask-RCNN for Mourmelon-le-Grand, showing the model's class accuracy (left), error rates for false negatives (center) and class accuracy for foreground proposals, as a function of the number of epochs (x-axis). These plots were monitored online via TensorBoard~\cite{tensorflow2015} in order to limit regularization issues.}
\end{centering}
\end{figure} 

\begin{figure}[!htbp]
\begin{centering}
\includegraphics[scale=0.2]{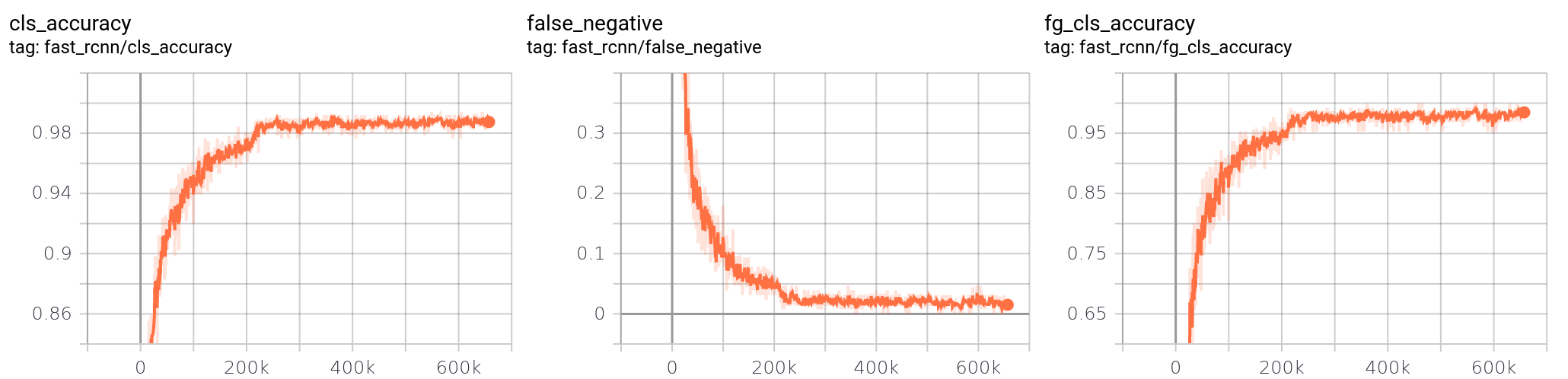}
\caption{\label{A4b} Training of the 3D reconstruction with Mask-RCNN for Sissonne, showing the model's class accuracy (left), error rates for false negatives (center) and class accuracy for foreground proposals, as a function of the number of epochs (x-axis). These plots were monitored online via TensorBoard~\cite{tensorflow2015} in order to limit regularization issues.}
\end{centering}
\end{figure} 

\begin{figure}[!htbp]
\begin{centering}
\includegraphics[scale=0.2]{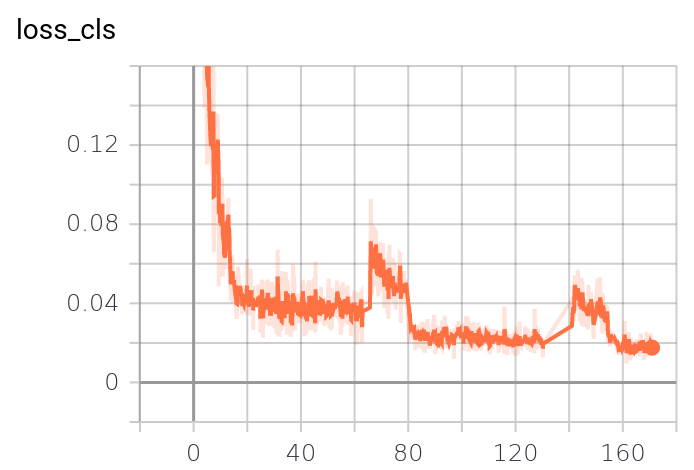}
\includegraphics[scale=0.2]{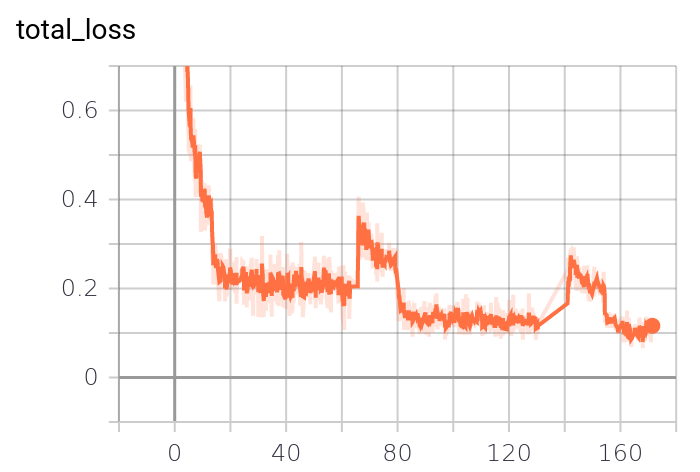}
\caption{\label{A5a} Training of the 3D reconstruction with Mask-RCNN for Mourmelon-le-Grand, showing the model's class loss (left) and total loss (right), as a function of the number of epochs (x-axis). These plots were monitored online via TensorBoard~\cite{tensorflow2015} in order to limit regularization issues.}
\end{centering}
\end{figure} 

\begin{figure}[!htbp]
\begin{centering}
\includegraphics[scale=0.2]{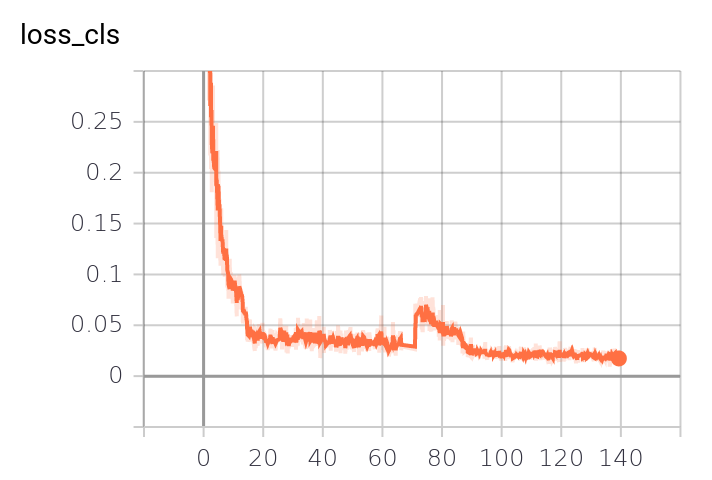}
\includegraphics[scale=0.2]{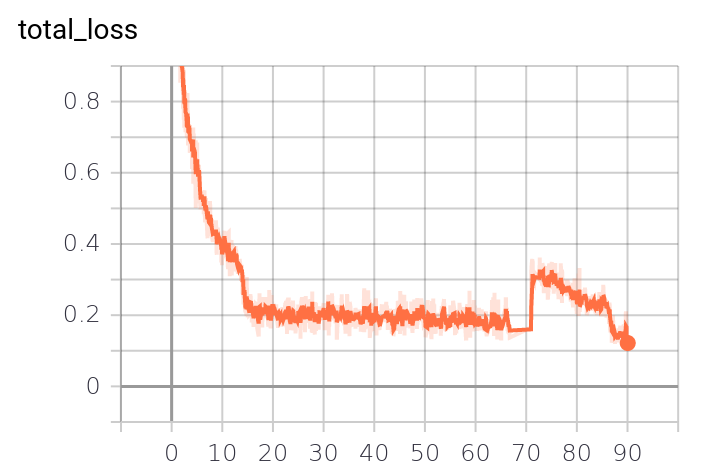}
\caption{\label{A5b} Training of the 3D reconstruction with Mask-RCNN for Sissonne, showing the model's class loss (left) and total loss (right), as a function of the number of epochs (x-axis). These plots were monitored online via TensorBoard~\cite{tensorflow2015} in order to limit regularization issues.}
\end{centering}
\end{figure}

\clearpage
\subsection{Ablation studies}
\label{SectionVId}

\begin{figure}[!htbp]
\begin{centering}
\includegraphics[scale=0.28]{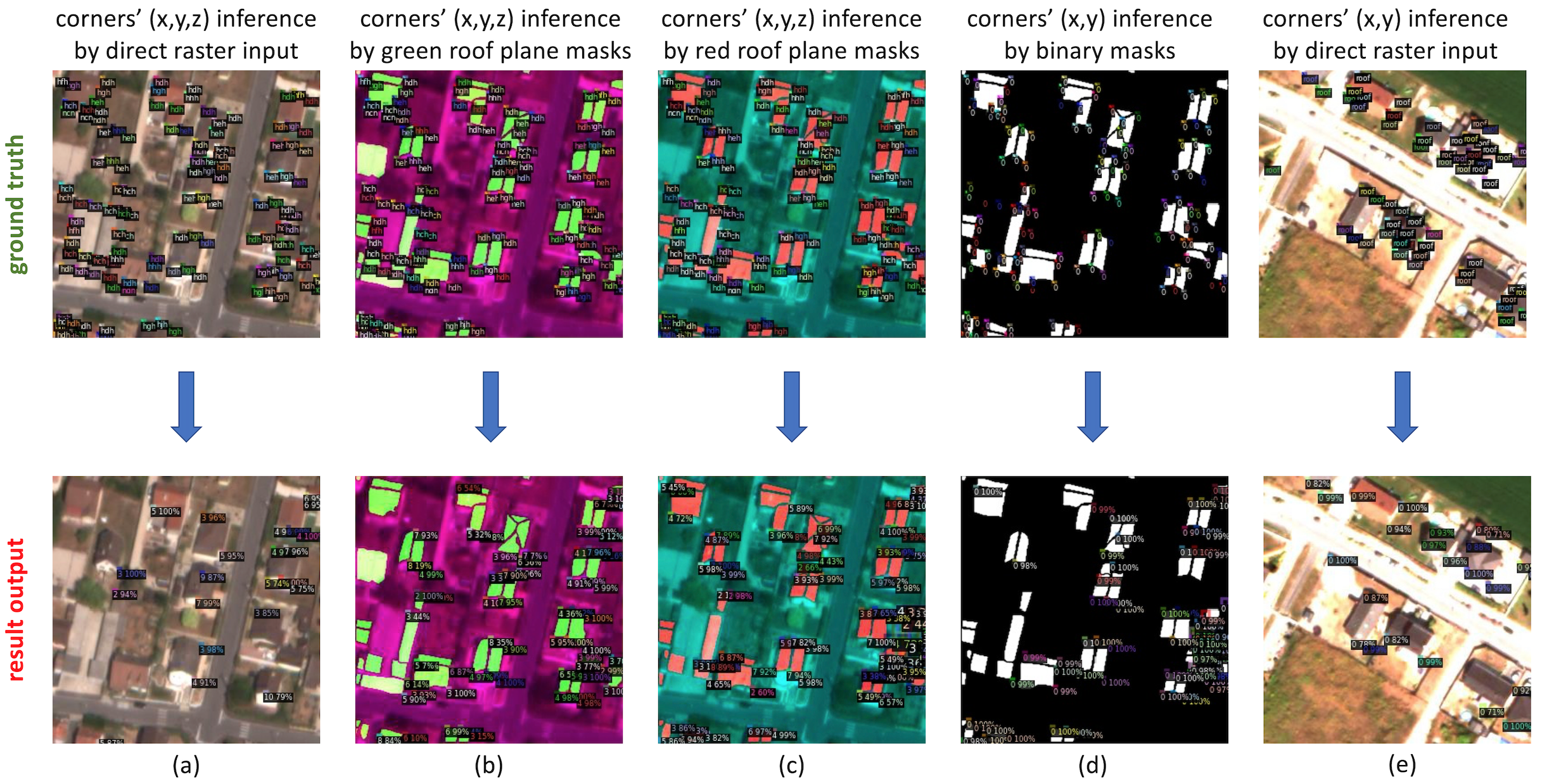}
\caption{\label{S1} Sum up of the ablation studies and other research approaches. If one tries to infer the roof section corners' position and heights $(x, y, z)$ directly through the raster satellite image (a), the result output shows a poor performance. In fact, even the simple inference of the roof section corners' positions $(x, y)$ is unsatisfying, whether the image input consists in the direct raster satellite image or the binary masks of the roof sections (d and e). Trying to perform the KIBS method through other RGB channels, e.g. green or red (b and c), doesn't change the performance of the algorithm noticeably.}
\end{centering}
\end{figure} 

\begin{figure}[!htbp]
\begin{centering}
\includegraphics[scale=0.3]{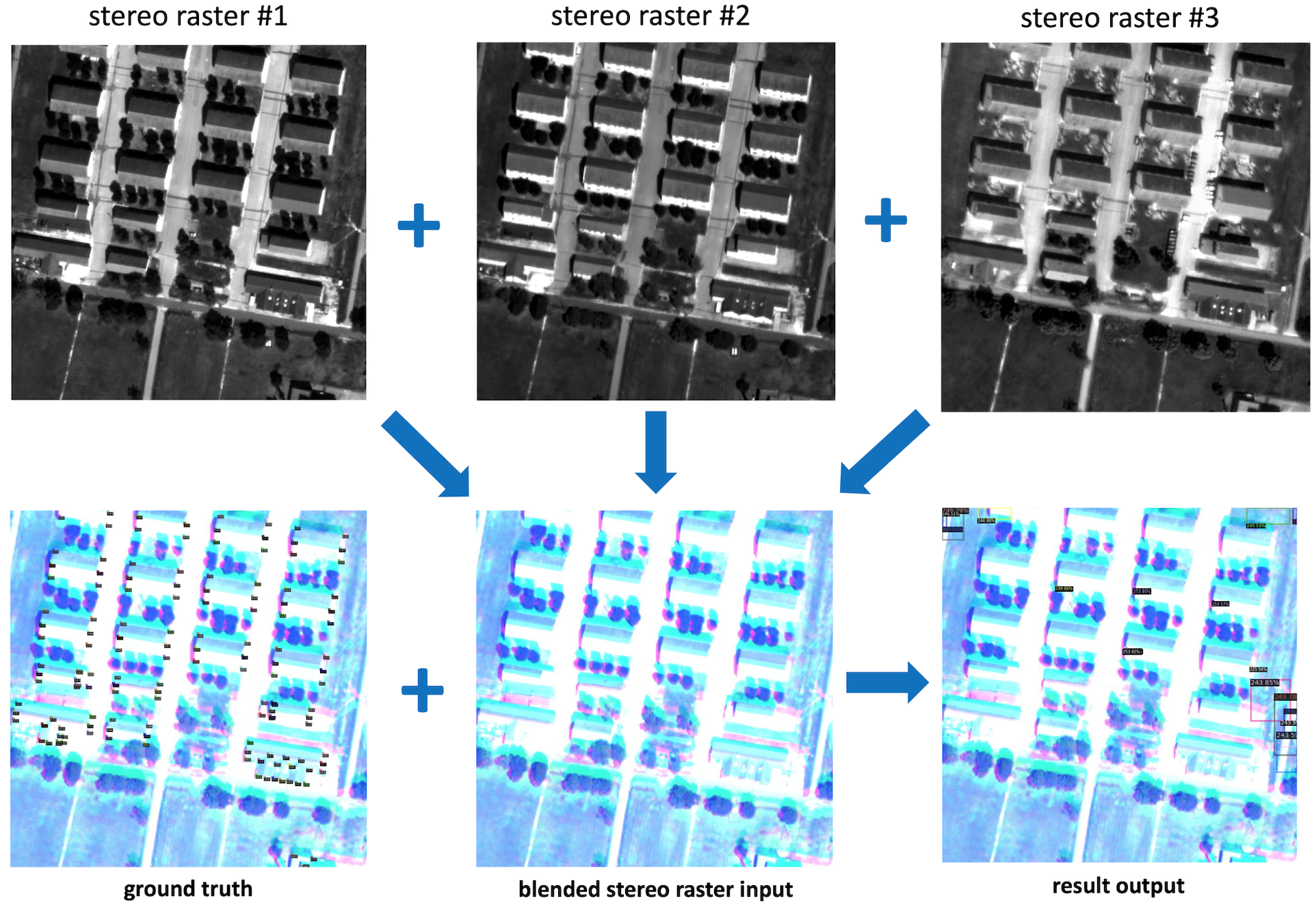}
\caption{\label{S2} Ablation study based on inferring the roof section corners' height out of a blending of three stereo raster pictures of the same geographical area, taken with different satellite viewing angles and solar zenith angles. As one can see, the output results show the poor performance of this approach.}
\end{centering}
\end{figure} 

\begin{figure}[!htbp]
\begin{centering}
\includegraphics[scale=0.09]{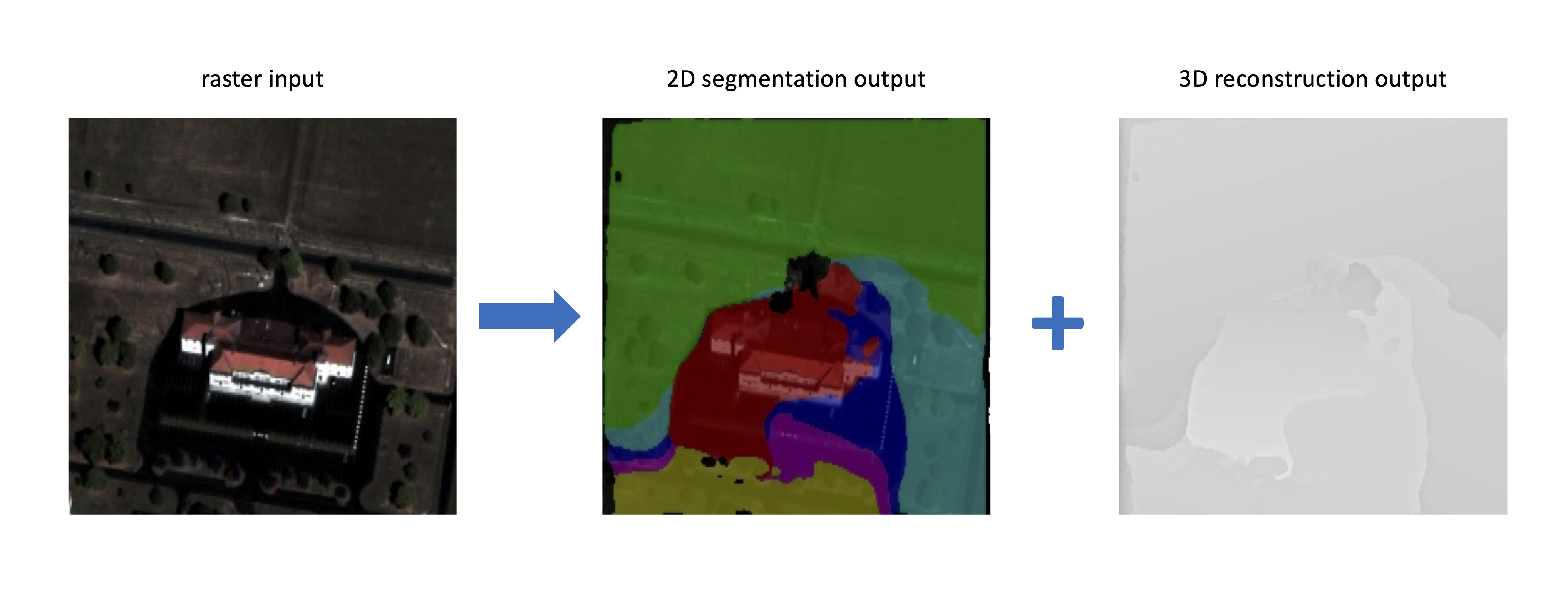}
\caption{\label{S3} Ablation study based on the work of ~\cite{Yu2019,Lenssen2020} instead of the Mask-RCNN algorithms used in the KIBS method, where a 2D segmentation is applied to the satellite raster input, and a 3D map is reconstructed out of it.}
\end{centering}
\end{figure} 

\clearpage
\subsection{Data postprocessing}
\label{SectionVIe}

There are six general comments one can make wrt. the data postprocessing of the KIBS method. We also refer the reader to Fig. \ref{A2} for a general recap of this procedure. 

Firstly, one needs to beware the procedure sometimes fails to correctly infer one or several roof section corners. For a number $N \geqslant 4$ of segmented roof section corners, the algorithm proceeds to select three corners among these forming the largest triangle area by a Delaunay triangulation, so as to increase 3D reconstruction accuracy. For $N=1$ or $N=2$, the model considers for simplification purposes the roof section to be parallel to the ground, and at a height equal to that of this corner, or the average of these two corners, respectively. Finally, if $N=0$ and no corner is detected by the algorithm, the roof section is also considered to be parallel to the ground, but assigned a height equal to the mean of all roof corners heights of the training set (which in the case of our data set amounts to $6.11$ m). 

Secondly, another basic postprocessing consists in ``filing'' the heights of the corners used for the roof section 3D reconstruction, based on the assumption that virtually no real roof section contains three corners of different heights. Let's assume these three corner heights $z_1, z_2, z_3$, in ascending order, are all unequal to each other: then, if $z_2 < (z_1+z_3)/2$, the algorithm hence sets $z_1$ and $z_2$ to the value of their average $(z_1+z_2)/2$; and if $z_2 \geqslant (z_1+z_3)/2$, the algorithm sets $z_2$ and $z_3$ to the value of their average $(z_2+z_3)/2$. 

Thirdly, the output of the this Mask-RCNN model for 3D reconstruction gives after some postprocessing (and some changes to the native Detectron2 code~\cite{biyologithub}), $15 \times 15$ pixels red squares representing the roof section corners, and the class of each of these corners (i.e. their height-to-ground in meters) is embedded in RGB format by assigning these pixels a value $200+z$ in the red channel, where $z \in \mathbb N^{\ast}$ is their height in meters (as shown in Fig. \ref{B2a} for Mourmelon-le-Grand and Fig. \ref{B2b} for Sissonne). This 3D reconstruction output of red squares over a black background is then blended over its associated 2D segmentation output (i.e. the blue roof sections over a black background). This blending process must be done cautiously for several reasons, and the KIBS code contains several postprocessing methods to ensure no data is lost or mismatched at this stage. The reasons are the following: i- in certain complex roof structures, some of these $15 \times 15$ corner squares can overlap other roof section segmentation pixels they don't belong to, and hence assign a wrong height to them; ii- some of these corner squares can sometimes be placed at inference sufficiently far away from the segmented roof section, so that no match is made and the whole roof section is ill-reconstructed; iii- the Delaunay triangulation will have to chose for each $15 \times 15$ pixels square only one pixel overlapping the segmented roof section rim, and hence a dedicated method must find and select this pixel among many others overlapping the plane. This data postprocessing pipeline of the 3D reconstruction output is shown in Fig. \ref{A3}. 

Fourthly, the segmented roof sections need to be perfectly distinguished (i.e. pixel-separated) from each other at the 3D reconstruction stage, since each has its own 3D plane coefficients inferred by the model. 

Fifthly, when these individual blended tile raster images are put back together to form the large $8687 \times 9890$ original image corresponding to the full satellite view, some roof sections and corners may be found to intersect two or more of these former tile images. Hence, some parts of a given reconstructed building may spread over several former tile images, and thus belong to different data sets (training, validation, or testing), with different associated blue pixel values. A \verb?flood_fill()? function from the scikit-image collection for image processing with right \verb?tolerance? parameter can correct this by assigning one single value via the blue channel of each spread roof section ($200, 210$, or $220$ for training, testing, or validation, respectively). This flood-fill is done on a first come, first served basis, with no particular priority from the training, validation, or testing set queues. 

Sixthly, the data postprocessing methods ultimately writes a text file containing for each line, each roof section points' 3D coordinates $\{x, y, z\}$ and ID ($0$ for training set origin, $1$ for validation set origin, $2$ for testing set origin). This is fed to the KSR method reconstruction~\cite{Yu2022} of the whole city in 3D for visualisation.

\section*{Acknowledgement}

We graciously thank LuxCarta for providing the satellite raster data with its hand-annotated ground truth. 

\newpage
\nocite{*}
\bibliographystyle{abbrvnat}
\bibliography{biblio}

\renewcommand{\theHsection}{\arabic{section}}
\appendix
\include{040_appendix}

\end{document}